\documentclass[sigconf]{acmart}

\usepackage[utf8]{inputenc} % allow utf-8 input
\usepackage{booktabs}       % professional-quality tables
\usepackage{amsfonts}       % blackboard math symbols
\usepackage{nicefrac}       % compact symbols for 1/2, etc.
\usepackage{microtype}      % microtypography
\usepackage{xcolor}         % colors
\usepackage{subcaption}
\usepackage{listings}
\usepackage{multirow}
\usepackage{here}
\usepackage[ruled,vlined,linesnumbered]{algorithm2e}
\usepackage{graphicx}
\usepackage{wrapfig}

\definecolor{dkgreen}{rgb}{0,0.6,0}
\definecolor{mauve}{rgb}{0.58,0,0.82}

\AtBeginDocument{%
  \providecommand\BibTeX{{%
    \normalfont B\kern-0.5em{\scshape i\kern-0.25em b}\kern-0.8em\TeX}}}

\copyrightyear{2024}
\acmYear{2024}
\setcopyright{acmlicensed}\acmConference[GECCO '24]{Genetic and Evolutionary Computation Conference}{July 14--18, 2024}{Melbourne, VIC, Australia}
\acmBooktitle{Genetic and Evolutionary Computation Conference (GECCO '24), July 14--18, 2024, Melbourne, VIC, Australia}
\acmDOI{10.1145/3638529.3654017}
\acmISBN{979-8-4007-0494-9/24/07}

\begin{document}

\title{LLMatic: Neural Architecture Search via Large Language Models and Quality Diversity Optimization}

\author{Muhammad U. Nasir}
\email{umairnasir1@students.wits.ac.za}
\affiliation{%
  \institution{University of the Witwatersrand}
  \city{Johannesburg}
  \country{South Africa}
}

\author{Sam Earle}
\email{se2161@nyu.edu}
\affiliation{%
  \institution{New York University}
  \city{New York}
  \country{USA}
}

\author{Christopher W. Cleghorn}
\email{christopher.cleghorn@wits.ac.za}
\affiliation{%
  \institution{University of the Witwatersrand}
  \city{Johannesburg}
  \country{South Africa}
}

\author{Steven James}
\email{steven.james@wits.ac.za}
\affiliation{%
  \institution{University of the Witwatersrand}
  \city{Johannesburg}
  \country{South Africa}
}

\author{Julian Togelius}
\email{julian@togelius.com}
\affiliation{%
  \institution{New York University}
  \city{New York}
  \country{USA}
}

% \author{
%     Muhammad U. Nasir$^{1,2}$,
%     Sam Earle$^{2}$, 
%     Julian Togelius$^{2}$,
%     Steven James$^{1}$, 
%     Christopher Cleghorn$^{1}$
% }
% \affiliation{%
% \country{umairnasir1@students.wits.ac.za, se2161@nyu.edu, julian@togelius.com, \\ 
%   steven.james@wits.ac.za, christopher.cleghorn@wits.ac.za}\\ 
%  \textsuperscript{\rm 1} \institution{University of the Witwatersrand}
%   %\department{School of Computer Science and Applied Mathematics}
%     \city{Johannesburg}
%   \country{South Africa}\\
%   \textsuperscript{\rm 2} \institution{New York University}
%   %\department{Tandon School of Engineering}
%     \city{New York}

%\author{Anonymous Authors}
\renewcommand{\shortauthors}{Muhammad U. Nasir, et al.}
%\renewcommand{\shortauthors}{Anonymous Author}

\begin{abstract}
  Large language models (LLMs) have emerged as powerful tools capable of accomplishing a broad spectrum of tasks. Their abilities span numerous areas, and one area where they have made a significant impact is in the domain of code generation. Here, we propose using the coding abilities of LLMs to introduce meaningful variations to code defining neural networks. Meanwhile, Quality-Diversity (QD) algorithms are known to discover diverse and robust solutions. By merging the code-generating abilities of LLMs with the diversity and robustness of QD solutions, we introduce \texttt{LLMatic}, a Neural Architecture Search (NAS) algorithm. While LLMs struggle to conduct NAS directly through prompts, \texttt{LLMatic} uses a procedural approach, leveraging QD for prompts and network architecture to create diverse and high-performing networks. We test \texttt{LLMatic} on the CIFAR-10 and NAS-bench-201 benchmarks, demonstrating that it can produce competitive networks while evaluating just $2,000$ candidates, even without prior knowledge of the benchmark domain or exposure to any previous top-performing models for the benchmark. The open-sourced code is available in \url{https://github.com/umair-nasir14/LLMatic}.
\end{abstract}

\begin{CCSXML}
<ccs2012>
   <concept>
       <concept_id>10010147.10010257.10010293.10010294</concept_id>
       <concept_desc>Computing methodologies~Neural networks</concept_desc>
       <concept_significance>500</concept_significance>
       </concept>
    <concept>
       <concept_id>10003752.10003809.10003716.10011136.10011797.10011799</concept_id>
       <concept_desc>Theory of computation~Evolutionary algorithms</concept_desc>
       <concept_significance>500</concept_significance>
       </concept>
   <concept>
       <concept_id>10010147.10010257.10010258.10010262.10010278</concept_id>
       <concept_desc>Computing methodologies~Lifelong machine learning</concept_desc>
       <concept_significance>300</concept_significance>
       </concept>
 </ccs2012>
\end{CCSXML}

\ccsdesc[500]{Computing methodologies~Neural networks}
\ccsdesc[300]{Computing methodologies~Lifelong machine learning}
\ccsdesc[500]{Theory of computation~Evolutionary algorithms}

\keywords{large language models, neural networks, quality-diversity optimization, neural architecture search}


\maketitle

\section{Introduction}

A major challenge in deep learning is designing good neural network architectures. Neural Architecture Search (NAS) is the generic term for various approaches to automating this design process~\cite{white2023neural}. The idea is to formulate an objective, such as maximum accuracy on a classification problem with a given budget of parameters and training cycles, and cast the problem as a search for the architecture that maximizes the objective. Every test consists of training the candidate network architecture using some form of gradient descent on the chosen benchmark dataset to measure its performance. This typically means that many thousands of architectures are tested and discarded in the process.

Two common algorithmic approaches to NAS are reinforcement learning and evolutionary computation. 
%They are used somewhat differently.
Reinforcement learning approaches to NAS~\citep{jaafra2019reinforcement} train a controller (typically another neural network) that outputs network architectures; these network architectures are tested and their performance is used as a reward signal. Evolutionary computation approaches to NAS~\citep{liu2021survey}, on the other hand, directly search the space of neural architectures. A population of architectures are kept, and their performance is used as a fitness score. Evolutionary NAS approaches are similar to neuroevolution, which has existed since the 1980s~\citep{tenorio1988self, miller1989designing}, and one might even see NAS as a form of neuroevolution. The main difference is that in NAS, the search process does not concern the parameters of the neural network, only its architecture.

One could argue that search by evolutionary computation or reinforcement learning is quite mindless and wasteful, given how many architectures need to be tested and how uninformed the changes that lead to each new architecture are. Is there some way we can inform the search by exploiting stored knowledge about how to design neural networks? This paper explores the idea that we can do exactly this using code-generating large language models (LLMs). More precisely, we propose combining an LLM with an evolutionary algorithm to generate new architectures that have high network architectural diversity and state-of-the-art performance.

The argument for this is simply that modern LLMs fine-tuned on code are very capable~\citep{shinn2023reflexion}. Given the amount of machine learning code they have been trained on, it is not surprising that they can design good neural network architectures. However, an LLM by itself cannot, in general, find an optimal architecture for a given problem, as it cannot test architectures and learn from its experiments. Therefore, we propose combining the domain knowledge of code-generating LLMs with a robust search mechanism.

While generating a single architecture that maximizes a given objective is useful for many cases, there is often more value to generating a set of architectures that vary across some relevant dimensions. For example, one might want to have a set of architectures that vary in their parameter counts or depths. This helps in understanding the trade-offs between various desirable metrics and could assist in making better-informed decisions about which architecture to use for a specific application. For example, one might want a range of networks for edge deployments to clients with different RAM sizes. To enable this, the solution we propose here leverages quality-diversity search~\cite{pugh2016quality}, specifically a version of the MAP-Elites algorithm~\citep{mouret2015illuminating}.

Our main contribution is a novel LLM-based NAS algorithm, \texttt{LLMatic}, that utilizes the power of two QD archives to search for competitive networks with just $2,000$ evaluations. We empirically show the performance of \texttt{LLMatic} on the CIFAR-10 dataset and the NAS-bench-201 benchmark where \texttt{LLMatic} searches for networks with performance similar to state-of-the-art results.

% \begin{itemize}

% \item Transformers \TODO{citation} started the era of Large Language Models (LLMs) \TODO{citation}. Apart from creating state-of-the-art models in core natural language processing tasks, they led to creating models for a wide variety of tasks, such as generating video game levels and code \TODO{citations}. 

% \item LLMs still exhibit limitations while generating complex tasks \cite{chen2021evaluating,todd2023level}.

% \item We introduce LLMatic, a Neural Architecture Search (NAS) algorithm to improve the neural network generating capabilities of an LLM.

% \item LLMatic uses the power of the Quality-Diversity (QD) algorithm to search for the best architecture. QD algorithms have previously been used for many complex tasks to find a range of diverse and high-quality solutions. \TODO{citations to a few QD papers.} 

% \end{itemize}
\section{Related Work}

% \begin{itemize}
    % \item \textbf{NAS}
    % \item \textbf{Code generating LLMs}
    % \item \textbf{QD}
    % \item EvoPrompting~\cite{chen2023evoprompting}
    % \item ML-Copilot~\cite{}
% \end{itemize}

Designing neural architectures can be an expensive and unintuitive process for human designers. Neural Architecture Search (NAS) aims to automatically find  architectures capable of strong performance after training~\citep{elsken2019neural}. 
% Naive search methods such as grid or random search can be used when the space of possible architectures is sufficiently small~\cite{}. 
Bayesian methods are a popular choice given their low sample complexity and the fact that evaluating each architecture (by training it) can be computationally expensive~\citep{kandasamy2018neural}. Alternatively, reinforcement learning can be used to train an agent (usually another neural network) to output candidate architectures for a given task, with the performance after training of the candidate architecture acting as a reward signal~\citep{jaafra2019reinforcement}. Evolutionary methods can also be used to search directly through the space of possible architectures~\citep{liu2021survey}. Similarly, Monte Carlo Tree Search has also been used to search~\citep{wistuba2017finding}. In all cases, a human designer must manually define a set of atomic network components or edit actions for use in network search/generation.

To avoid having the designer constrain the space of possible architectures prior to search, we turn to code-generating large language models (LLMs)---large models trained auto-regressively on massive datasets of code (e.g. public repositories hosted on Github). 
% Transformers~\citep{vaswani2017attention} facilitated the explosion of LLMs~\citep{radford2019language, brown2020language}. Apart from creating state-of-the-art models in core natural language processing tasks~\citep{adelani2022few, nasir2022geographical}, they led to creating models for a wide variety of other tasks, such as generating video game levels and code~\citep{chen2021evaluating, todd2023level, nasir2023practical}. 
These LLMs are based on the transformer architecture~\citep{vaswani2017attention} that has obtained state-of-the-art performance in natural language modelling~\citep{radford2019language,brown2020language,adelani2022few,nasir2022geographical}. They have also been successfully used in specific applications, such as for video game level design~\cite{todd2023level,nasir2023practical,chen2021evaluating} or code generation~\citep{shinn2023reflexion}.

Recently, LLMs have been used for evolving code by framing code-generation as an evolutionary problem. Evolution through Large Models (ELM)~\citep{lehman2022evolution} casts LLMs as evolutionary operators within a MAP-Elites~\citep{mouret2015illuminating} algorithm tasked with evolving a robot's morphology at code-level. EvoPrompting~\citep{chen2023evoprompting} is an LLM-based method that is somewhat similar to ours in that it uses code-LLMs as mutation and crossover operators to perform NAS. It is tested on the MNIST-1D classification task~\citep{greydanus2020scaling} and the CLRS algorithmic reasoning benchmark~\citep{velivckovic2022clrs}. Since performance can generally be trivially increased by simply adding parameters to the model, an additional penalty is added to the fitness of a candidate neural architecture corresponding to its model size. This incentivizes the discovery of small models with effective architectures. In our method, we instead consider model complexity (in terms of FLOPS) as a diversity metric, searching for high-performing models of a variety of sizes. GENIUS~\citep{zheng2023can} is another LLM-based NAS algorithm that uses GPT-4 to simply search through straight-forward prompting.

Quality Diversity (QD) methods~\citep{pugh2016quality} are a family of evolutionary algorithms that, in addition to optimizing a fitness metric, search for a diversity of individuals according to some user-specified ``behavioral descriptors''. Instead of keeping a population of the fittest individuals, QD methods such as MAP-Elites~\citep{mouret2015illuminating} maintain an ``archive'' of individuals, where this archive is partitioned into cells, with each cell corresponding to individuals exhibiting a particular range of values along each behavioral descriptor.

QD methods are valuable in domains such as robot control, where it is useful to learn diverse high-quality trajectories, in case one solution should become unavailable during deployment because of a physical obstruction or mechanical malfunction~\citep{cully2015robots}. Another motivating factor is that greedily searching for the fittest individual may not be desirable in deceptive domains. Here, maintaining a diversity of fit individuals may protect the population from falling into local optima~\citep{gaier2019quality}. Conversely, diverse, unorthodox solutions may provide valuable ``stepping stones'' on the path to globally fit individuals.

\begin{figure*}[!htb]%{r}{0.55\textwidth}
    \centering
    \includegraphics[width=0.75\textwidth]{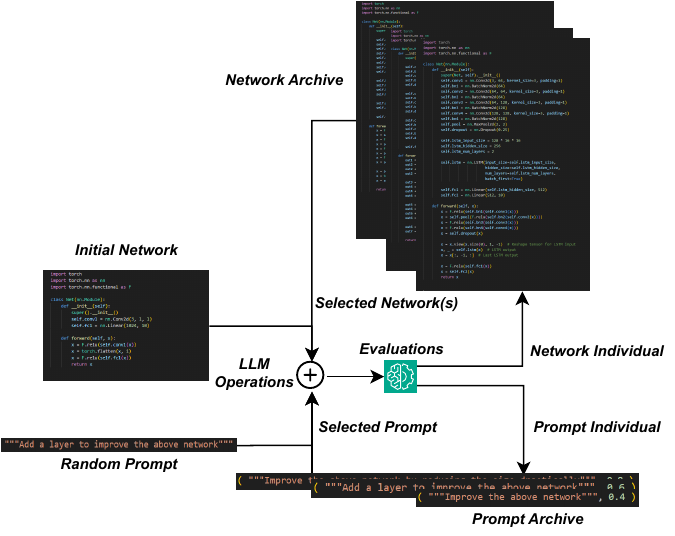}
  \caption{Illustrated in the figure is the flow of LLMatic. In the initial round of evolution, an initial network with a random prompt goes through a mutation operation. Network individual and prompt individual are then evaluated to be stored in separate archives. During the evolutionary loop, the selected prompt and network go through an evolutionary operation (the prompt is fixed if the operation is crossover) to create more networks and prompt individuals to fill and illuminate the archives.}
    \label{fig:flowchart}
\end{figure*}

\section{Approach}

%\texttt{LLMatic} harnesses the power of LLMs and QD optimization to generate neural networks that are competitive with minimal searches. 

%\begin{figure}[!ht]
%    \centering
%    \includegraphics[width=0.8\linewidth]{Figures/LLMatic.pdf}
%  % \caption{\TODO{Make a better version of it.}}
%    \caption{An overview of the LLMatic pipeline. Two QD archives are maintained throughout %evolutionary search: one for networks of varying depth-to-width and overall complexity, and one for %different text prompts and temperature pairings used to prompt the LLM to generate candidate neural %architectures.}
%    \label{fig:overview}
%\end{figure}

\texttt{LLMatic} begins its search with a very basic neural network, inspired by the work of ~\cite{stanley2002evolving} which suggests that neuroevolution tends to perform better when starting with a small network. %An extensive ablation study has been conducted for this. 
In \texttt{LLMatic}, we use a novel dual-archive cooperative QD optimization approach, in which two separate archives are used to store complementary components that can be combined to solve a given task. The first archive stores neural networks, where the width-to-depth ratio and Floating Point Operations per Second (FLOPS) of a network are the behavioural descriptors. The width-to-depth ratio is a division of the width and the depth of the network. To specify the width, we use the maximum of the output features of all layers, while depth is simply the number of layers. Note that we choose FLOPS instead of parameter count because FLOPS correlates better with actual time spent training a network \cite{ayala2017efficient}. %two networks with the same number of parameters can have different numbers of FLOPS; counting FLOPS therefore allows more diversity. 
We call this archive the ``network archive''. The fitness function for the networks in this archive is defined as the test accuracy of the network after training. The second archive, called the ``prompt archive'', contains the prompt and temperature used for generating code, which are also the behavioural descriptors. The \textit{temperature} of an LLM is a hyperparameter that controls the degree of stochasticity in the selection of output tokens: a lower temperature means that the LLM will select tokens more deterministically, while a higher value will result in more diverse output. 
% creativity of the LLM. Lower temperature means the LLM will not diverge from the answer, higher means it will be more creative. 
The selection of prompt and temperature depends on a curiosity score~\citep{cully2017quality}, governed by whether the generated network was added to the network archive. The fitness of prompt individuals depends on whether the network was better than the previous generation's best score. \autoref{fig:flowchart} illustrates the flowchart of the approach LLMatic uses, while Algorithm~\ref{algorithm:llmaticalgo} shows the complete search process of \texttt{LLMatic} in pseudocode.

%\begin{wrapfigure}{r}{0.7\textwidth}
%    \centering
%    \includegraphics[width=1\linewidth]{Figures/algo.png}
%  %\caption{The illustration of the best accuracy per generation for %\texttt{LLMatic} and all ablation studies. Each experiment is %conducted with 10 seeds. The shaded region is the standard deviation %while the solid line represents the mean. EfficientNet-B0 is the best-%performing EfficientNet on CIFAR-10.} %Each experiment is observed %with the same hyperparameters as tuned for LLMatic.}
%    \label{fig:lossGen}
%\end{wrapfigure}

In the first generation, a simple neural network with one convolutional and one fully connected layer initiates the evolution (line 1 of Algorithm \ref{algorithm:llmaticalgo}). A prompt is selected at random to generate an initial batch of networks (lines 5--6 of Algorithm \ref{algorithm:llmaticalgo}). These networks are evaluated and an attempt is made to add them to the network archive as a random initialization for MAP-Elites. Concurrently, we mutate the temperature based on the fitness of the network, increasing it if the fitness increases and vice versa (lines 21--23 of Algorithm \ref{algorithm:llmaticalgo}).
An increase in temperature is desirable if we want the LLM to explore, while decreasing the temperature will result in the LLM exploiting in an attempt to achieve better fitness than before. Once we calculate the fitness of the prompt individual, we add the score to a collective prompt fitness score, after which we try to populate the prompt archive. The collective prompt fitness score determines the overall fitness of each individual in the prompt archive as it gives each prompt a fitness score.

Once either of the archives reaches a specified capacity, we introduce neural network training and evolutionary operators in the process (lines 7--20 of Algorithm \ref{algorithm:llmaticalgo}). With a certain probability at each generation, a decision is made on whether to perform crossover or mutation to produce $N$ new offspring. If the crossover operator is chosen, we select $N$ random network individuals, locate their closest networks in the archive, and carry out a crossover operation instructed by a prompt (lines 15--16 of Algorithm \ref{algorithm:llmaticalgo}). No individual is added to the prompt archive when a crossover is performed. If the mutation operation is selected, we pick the most curious prompt individual and a random network individual. For exploration, we also select random prompts. In both cases, each network is trained for a certain number of epochs and an attempt is made to add the network to the archive. Likewise, a prompt individual is added as previously described. This process continues for a predetermined number of generations. Refer to the supplementary material for pseudocode on mutation operators, crossover operators, temperature mutation and addition to archives.

\begin{algorithm}[h!]
\
\SetAlgoLined
\SetKwInOut{KwIn}{initialize}
\SetKwInOut{KwOut}{Output}
\SetKwFor{For}{for}{}{}%
\SetKwFor{While}{while}{}{end}
%\KwIn{network\_archive, prompt\_archive, best\_loss, initial\_network;}
%selected\_network = initial\_network;
Initialize network and prompt archives.

\While{number of generations $<$ maximum generations}{

\ForEach{network in batch of networks}{

\eIf{number of individuals in archives $<$ set threshold}{
   
    %selected\_prompt = rand\_prompt;
        
    %generated\_network, temperature= \texttt{\textit{Mutation}};
        
    %prompt\_individual = selected\_prompt, temperature;
        
    %add\_to\_archive(selected\_net, prompt\_ind, best\_loss);
    %\texttt{\textit{add\_to\_archive}};
    \texttt{\textit{Mutate the initial network with a random prompt}}.
        
    Get the network and prompt individuals and add them to the respective archives. 
    
}{
    Randomly choose mutation or crossover as the evolutionary operator.

        \eIf{evolutionary operator $==$ mutation}{
        
            Select network and prompt.
            
            %rand\_network\_from\_network\_archive;

            %from\_prompt\_archive\_get\_           half\_curious\_prompt\_individuals\_            half\_random\_prompts\_individuals;
            
            %generated network, temperature = \texttt{\textit{Mutation(Selected network and prompt)}};
            \texttt{\textit{Mutate the selected network with the selected prompt}}.
            %prompt individual = selected prompt, temperature;
            %            all\_networks += generated\_network;
            
            Train or query the generated network.
            
            %all\_prompts += prompt\_individual;
            Get the network and prompt individuals and store them. 
            }{
            
            Selection of networks for crossover.
            %generated\_network = \texttt{\textit{Crossover}};
            
            \texttt{\textit{Perform crossover on selected networks with a fixed prompt}}.
            %all\_networks += generated\_network;
            
            Train or query the generated network.
            
            Store the generated network individual.
        }
}

}

Evaluate all networks to find the test accuracies.

\texttt{\textit{Mutate the temperature for the next generation}}.

Add the batch network individuals and the corresponding prompt individuals in the respective archives.

%\texttt{\textit{add\_to\_archive}};

}

 \caption{LLMatic}

 \label{algorithm:llmaticalgo}
\end{algorithm}

\section{Evaluating LLMatic}

% \subsection{Dataset}

To evaluate \texttt{LLMatic}, we use CIFAR-10~\citep{krizhevsky2009learning}, a commonly used dataset for NAS \citep{tan2019efficientnet, ying2019bench}. We perform extensive ablation studies to demonstrate that \texttt{LLMatic} benefits from each of its components during search. Once our algorithm is validated, we extend our experiments to \emph{NAS-bench-201}. The \emph{NAS-bench-201} benchmark~\cite{mehta2022bench} is a dataset enumerating all possible neural architectures within a given search space (i.e. of a fixed number of nodes/layers and edges/operations) and the corresponding test accuracy of each architecture after training on a given dataset (e.g. CIFAR-10), allowing researchers to explore the tradeoffs of various NAS algorithms without needing to retrain each candidate network during search.

\subsection{Setting up LLMatic}

\textbf{Dataset:} The CIFAR-10 dataset is made up of $60,000$ color images, each with a resolution of $32 \times 32$ pixels, and divided into 10 categories: airplane, automobile, bird, cat, deer, dog, frog, horse, ship and truck. %Each category contains 6,000 images. Out of the total images, 50,000 are designated for training and 10,000 are set aside for testing.
The dataset is partitioned into five groups for training and one group for testing, each group holding 10,000 images. Each test group consists of an exact count of 1,000 images from each category selected randomly. The training groups hold the remaining images, which are arranged in random order. As a result, some training groups might contain more images from one category compared to others. Nonetheless, collectively, the training groups have an exact total of 5,000 images from each category.

\textbf{Initial Neural Network:} \texttt{LLMatic} starts off with a simple neural network with one convolutional layer that takes in $3$ input channels, with $1 \times 1$ kernel size and $1$ output channel which connects to a dense layer of size $1024$. 
% Since the size of images in the dataset is $32 \times 32$, with $1$ (grayscale) channel, the linear layer will have $1024$ hidden neurons.
These hidden neurons are connected via another dense layer to $10$ output neurons (as we have 10 classes). Rectified Linear Unit (ReLU)~\citep{nair2010rectified} is the activation function used in all layers. All of our networks are generated in PyTorch~\citep{paszke2019pytorch}.

\textbf{Generating Neural Networks:} At each generation, we generate a batch of $100$ new offspring. Each network generated is trained for $50$ epochs. The networks are optimized by stochastic gradient descent~\citep{bottou2010large} with the learning rate set to $0.001$ and momentum set at $0.9$ for all networks. We use cross entropy loss as our measure for the fitness of the trained network. 

For evolutionary operators, we set a probability of $0.7$ for mutation and $0.3$ for crossover as after experimentation, we found that mutation creates consistently more trainable neural networks. We initialize the temperature parameter (used when sampling the code-generating LLM) to $0.6$. For temperature mutation, half of the population is generated by the prompt individual temperature mutated uniformly at random between $-0.1$ to $0.1$. The other half is generated by the temperature obtained from the prompt individual itself. If the fitness of the generated network is better than or equal to the best fitness of the previous generation, we increase the temperature by $0.05$ and if it is worse than the best fitness of the previous generation, we decrease it by $0.05$. For the crossover operator, we select $10$ random networks and find their $2$ or $3$ nearest neighbours, based on the distance of niches, in the network archive to perform crossover. We set the LLM temperature to be $0.7$ for network generation.

\textbf{Quality Diversity Optimization:} For our QD optimization algorithm, we choose a variant of MAP-Elites---Centroidal Voronoi Tessellation (CVT-MAP-Elites) \citep{vassiliades2017using}---which can be seen as a generalization of MAP-Elites that is  intended to scale MAP-Elites to high-dimensional behavior spaces; CVT-MAP-Elites outperforms standard MAP-Elites in high-dimensional spaces while matching performance lower-dimensional scenarios such as those studied in the present work. 
% and is shown to do so by comparing with other variants of MAP-Elites by \citet{nilsson2021policy}.

CVT-MAP-Elites automates the sub-division of the archive by identifying $k$ cell centroid locations exhibiting an even spread through the behavioural descriptors space. Here, $k$ corresponds to the total number of evolutionary ``niches'' in the archive. We use the \textit{pymap\_elites}\footnote{\url{https://github.com/resibots/pymap_elites}} implementation for our experimentation. We use a k-d tree~\citep{bentley1975multidimensional} to create and write centroids to the archive and find the nearest neighbors using a Euclidean distance metric \citep{dokmanic2015euclidean}.

% For our QD archives, we use $10$ niches per dimension, and we have $2$ dimensions per archive.
Each of our QD archives has $2$ dimensions (behavioral descriptors), with $100$ niches spread across them. We set the number of random initial networks to $10$. 
% Random initial networks are needed to be filled in archives before evolutionary operators are introduced. 
For the network archive, we have the width-to-depth ratio of the network as our first dimension and the FLOPS  of the network as the second dimension. The width-to-depth ratio has a lower limit of $0$ and an upper limit of $200$. The minimum FLOPS is set to $200$ MegaFLOPS and the maximum is set to $5$ GigaFLOPS. This range is set after experimentation.

For the prompt archive, we have the prompt encoded as an integer as the first dimension and temperature as the second dimension. The maximum value of the prompt is 16, the number of prompts used in the system. The maximum temperature value is set to $1$ as it can never increase beyond that for our LLM. The lower limit for all dimensions is $0$. 

For the network archive, we simply select a random network while for the prompt archive, we select the most curious prompt individual, which depends on the curiosity score. This curiosity score is incremented by $1.0$ if the selected prompt adds the generated network to the network archive, decreased by $0.5$ if the network is not added, and reduced by $1.0$ if the created network is untrainable. If the generated network has better fitness than the previous generation's best network, the collective prompt fitness score for the prompt in the prompt individual is increased by $1$; otherwise, it is unchanged. We use prompts that are generalizable to any problem in any domain. Refer to supplementary material for an example of mutation and crossover prompts.

\textbf{Code Generating LLM:} We use the pre-trained CodeGen \citep{nijkamp2022codegen} LLM to generate neural networks. CodeGen is an autoregressive decoder-only transformer with left-to-right causal masking. CodeGen is first trained on \texttt{ThePile} dataset with random initialization and is called CodeGen-NL. CodeGen-Multi is initialized with CodeGen-NL and is trained on \texttt{BigQuery} dataset. Lastly, CodeGen-Mono is initialized with CodeGen-Multi and is trained on \texttt{BigPython}. CodeGen is trained to be in various parameter sizes, but we use $6.1$ Billion parameter variant of CodeGen-Mono due to computational constraints.

\texttt{ThePile} dataset \citep{gao2020pile} is an $825.18$ GB English text corpus. CodeGen selects a subset of the Google \texttt{BigQuery} dataset which contains 6 programming languages, namely C, C++, Go, Java, JavaScript, and Python. The authors collected a large amount of permissively licensed Python code from GitHub in October $2021$, and named it \texttt{BigPython}. The size of \texttt{BigPython} is $217.3$ GB.

CodeGen-6B has $33$ layers and $16$ heads with $256$ dimensions per head. The context length is $2048$ and the batch size is $2$ million tokens. Weight decay is set to $0.1$. $0.4e^{-4}$ is the learning rate. Warm-up steps are set to $3k$ while total steps for training are $150$k.

%We use Euclidean distance as the metrics to find nearest neighbours.

\subsection{Ablation Study}
%TODO: Add GA-based evolution. Would be a nice ablation study. Recommended by AAAI reviewer.
As we have many components in \texttt{LLMatic}, we choose to do a thorough ablation study to determine the effect of each component on overall performance. The following are the components tested for the ablation study:

\begin{itemize}
    \item \texttt{Network-Archive-LLMatic}: \texttt{LLMatic} with only the network archive. To achieve this, we create a population of prompt individuals. The population is fixed to $100$ individuals initialized with random individuals. We have only one fitness score for this population, which is calculated as $+1$ if a network is added in the network archive, $-0.5$ if the network is not added and $-1$ if the network is not trainable. After we generate the network, we mutate the temperature by adding $0.1$ if the network is added in the network archive and $-0.1$ if the network is not added.
    \item \texttt{Prompt-Archive-LLMatic}: \texttt{LLMatic} with only the prompt archive. To achieve this, we create a population of networks. The fitness function for the population of networks is accuracy. We keep the population to $100$ individuals. With a similar probability as \texttt{LLMatic}, we select mutation or crossover operator. For the crossover operator, we select the individual that is closest to the structure of the selected network. For network similarity, we use cosine similarity and we choose the networks with higher scores. For the mutation operator, similar to \texttt{LLMatic} we mutate half of the networks from the most curious prompt individuals and half from random individuals.
    \item \texttt{Mutation-Only-LLMatic}: \texttt{LLMatic} using only mutation.
    \item \texttt{Crossover-Only-LLMatic}: \texttt{LLMatic} using only crossover.
    \item \texttt{Random-NN-Generation}: Neural network generation without evolution. We generate $100$ networks per generation for $20$ generations as a fair comparison to LLMatic, which generates the same number per batch. We apply the prompt ``Create a neural network that inherits from nn.Module and performs better than the above neural network'' and we add the initial network with this prompt.
\end{itemize}

\begin{figure*}[h!]
    \centering
    \includegraphics[width=0.65\linewidth]{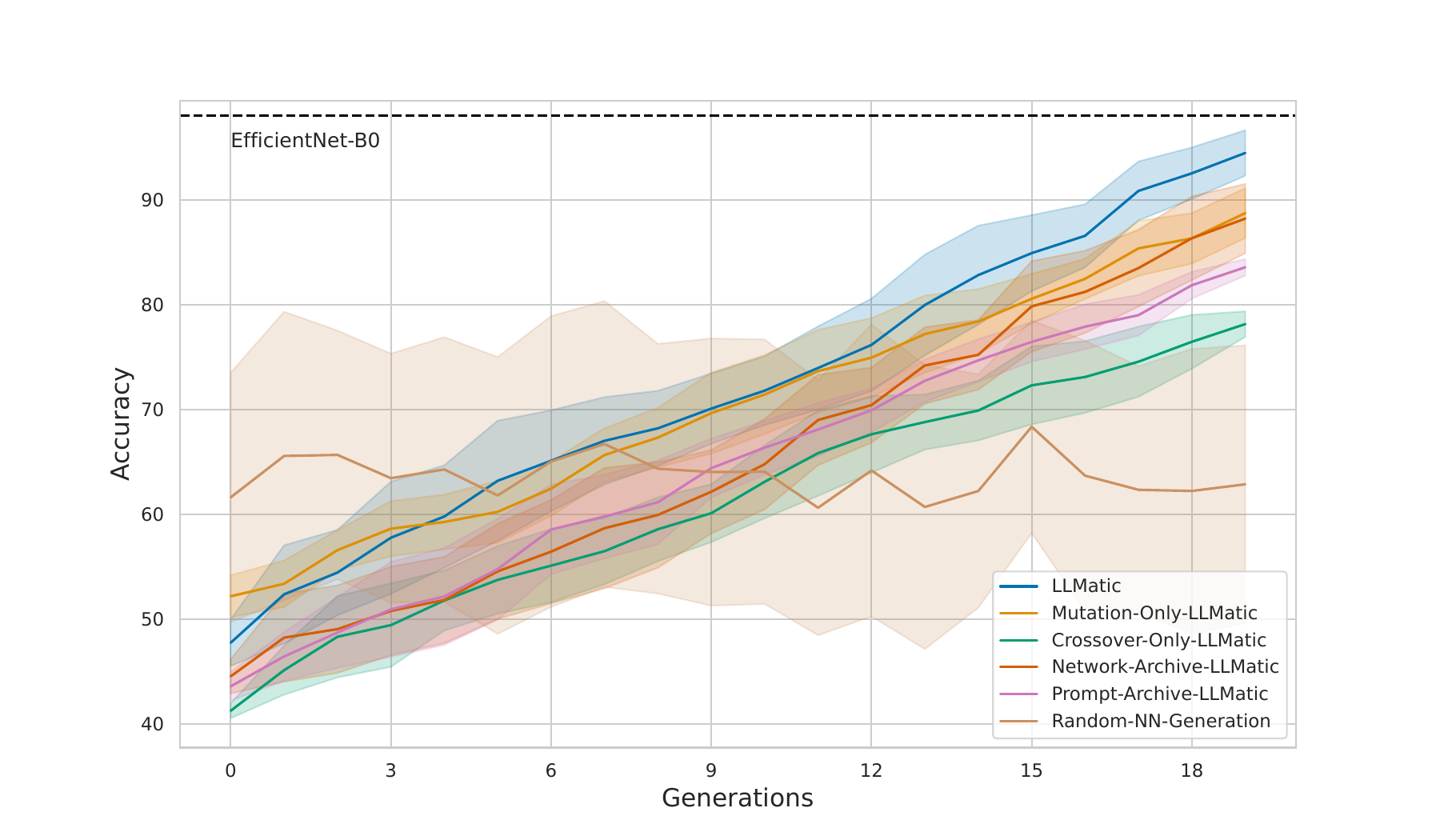}
  \caption{The illustration of the best accuracy per generation for \texttt{LLMatic} and all ablation studies. Each experiment is conducted with 30 seeds. The shaded region is the standard deviation while the solid line represents the mean. EfficientNet-B0 is the best-performing EfficientNet on CIFAR-10.} %Each experiment is observed with the same hyperparameters as tuned for LLMatic.}
    \label{fig:lossGen}
\end{figure*}

\newcommand{\www}[1]{0.49\linewidth}
\begin{figure}%{r}{0.5\textwidth}
    
    \centering
    \begin{subfigure}{\www}
        \includegraphics[width=1\textwidth]{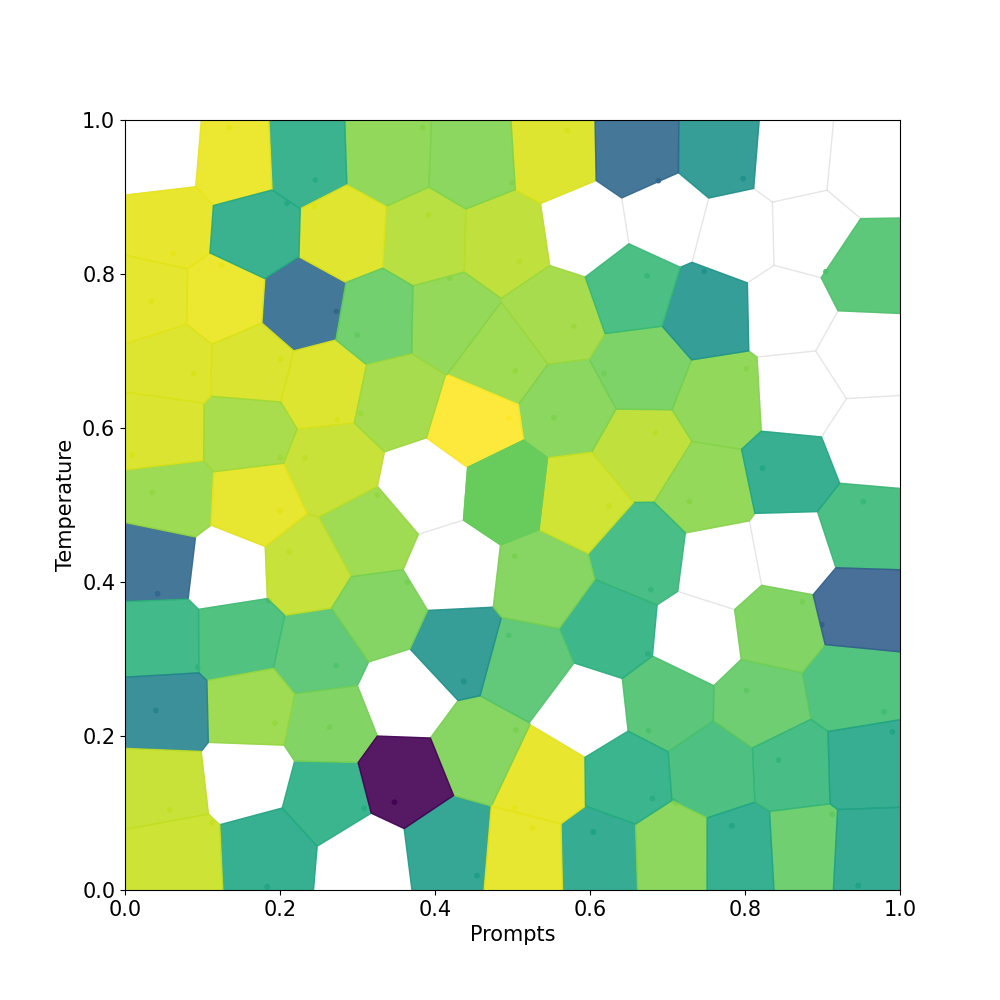}
        \caption{Prompt archive: Prompts encoded as integers on the $x$-axis, normalised to be in range 0-1 for CVT-MAP-Elites as all points are within $0$--$1$. On the $y$-axis, we have the temperature that controls LLMs exploration ability. As 1 is the maximum temperature, there is no need for normalisation.}
        \label{parch}
    \end{subfigure}
    \hfill
    \begin{subfigure}{\www}
        \includegraphics[width=1\textwidth]{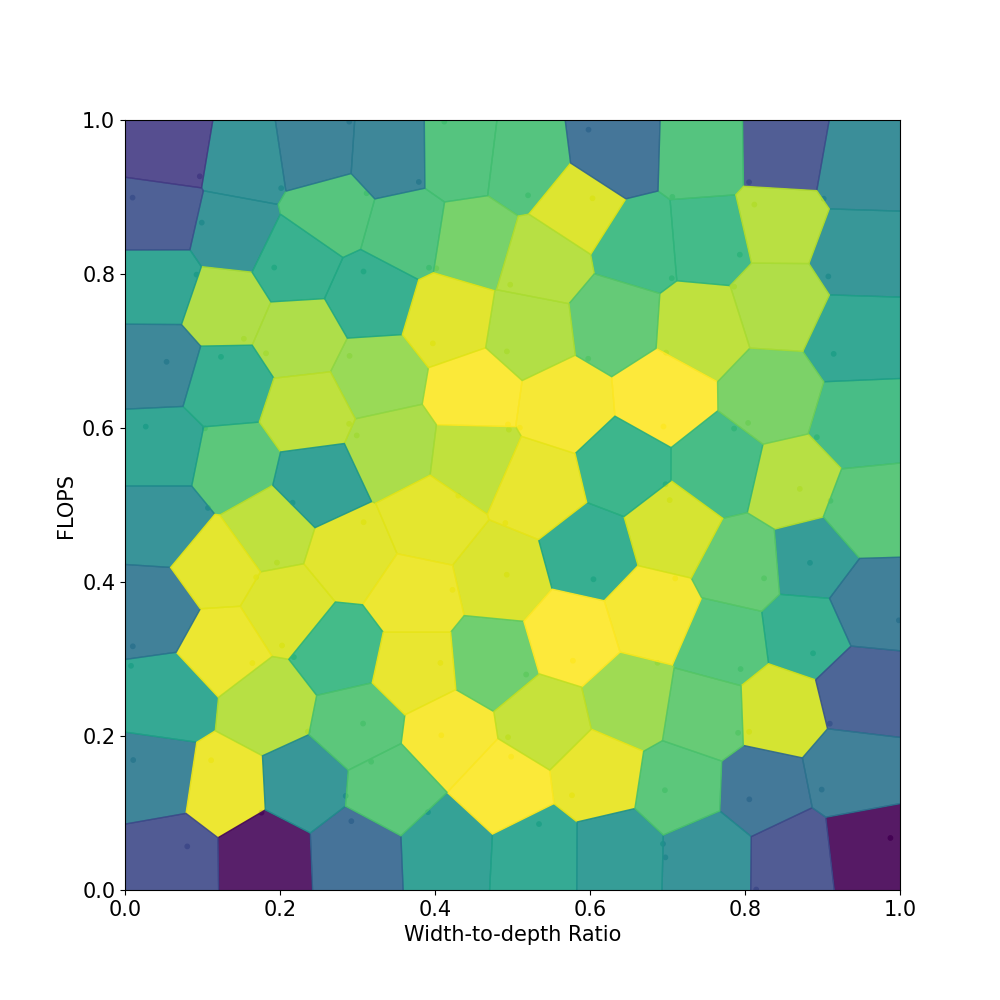}
        \caption{Network archive: Width-to-depth ratio on the $x$-axis. The range for the Width-To-Depth ratio is from $0$--$200$ normalised to $0$--$1$. On the $y$-axis, we have Floating Point Operations per Second (FLOPS). We have a range of 200 Mega FLOPS to 5Giga FLOPS. This range is normalised to 0-1 for CVT-MAP-Elites.}
        \label{narch}
    \end{subfigure}
    
    \caption{An illustration of archives generated by \texttt{LLMatic}. We have selected the archive with the median number of cells filled in experiments over 30 seeds. \autoref{parch} shows the prompt archive, while \autoref{narch} shows the network archive. The lighter the colour of the filled cell, the better fitness of the individual. White indicates that the cell is empty.}
    \label{fig:archives}
\end{figure}

\subsubsection{Ablation Results and Discussion}

\begin{figure}[h!]%{r}{0.5\textwidth}
    \centering
    \includegraphics[width=1.0\linewidth]{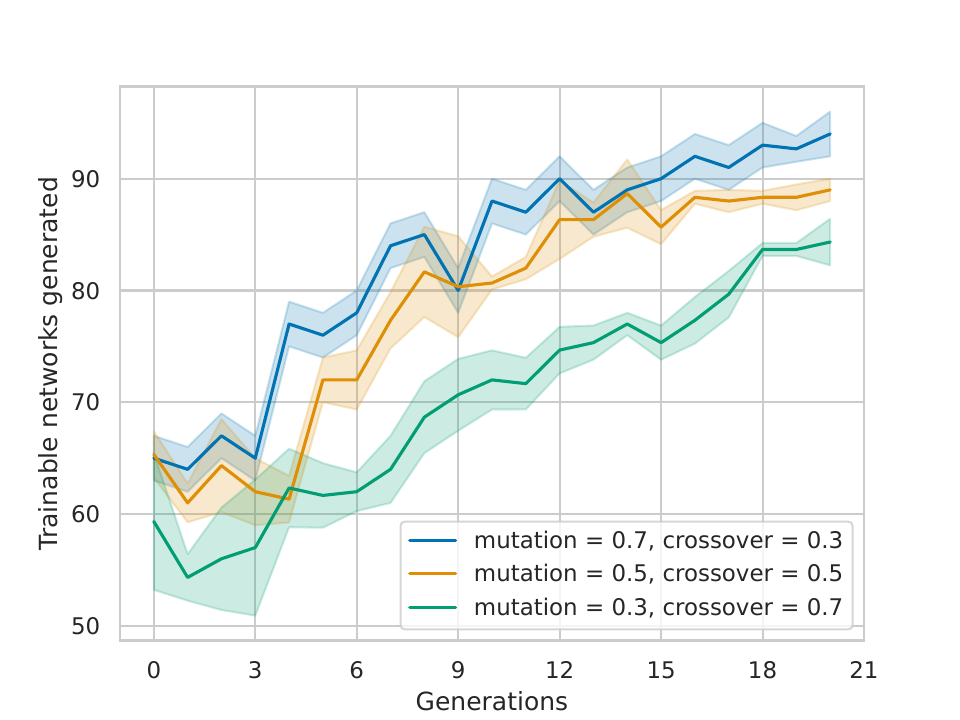}
  \caption{The illustration of how many trainable networks are created in a generation. The total number of networks created is 100 per generation. This illustration is calculated over 10 runs. The shaded region is the standard deviation. }
    \label{fig:crossmut}
\end{figure}

In this section, we will discuss the results of the experiments that we set up in the previous section. We first discuss the best accuracy per generation, illustrated in Figure \ref{fig:lossGen}. This will lead our discussion to trainable networks generated by changing the crossover and mutation probabilities (Figure \ref{fig:crossmut}). Then we will discuss how archives are illuminated Figure \ref{fig:archives}. Some of the generated networks are shown in the supplementary material.

Figure \ref{fig:lossGen} illustrates that each component of \texttt{LLMatic} is necessary. \texttt{Mutation-Only-LLMatic} and \texttt{Network-Archive-LLMatic} are the closest to \texttt{LLMatic}, which validates our choice to weight the probability of mutation higher. \texttt{Crossover-Only-LLMatic} performs the worst, as it does not benefit from the exploration abilities provided by the mutation operator \citep{ullah2022analysis}. Both operators (mutation and crossover) together provide exploration and exploitation abilities to \texttt{LLMatic}, which appear necessary to find high-quality and diverse networks. \texttt{Prompt-Archive-LLMatic} performs poorly, indicating that the network archive is an important aspect in finding high-performing networks. However, both archives together demonstrate competitive results.

We use EfficientNet-B0, which is the state-of-the-art network on CIFAR-10 \citet{tan2019efficientnet} as an indicator of where our algorithm stands. EfficientNet-B0 was searched via methods applied by \citet{tan2019mnasnet} and is slightly larger than the original study as they were targeting more FLOPS. 
The original study required $8,000$ evaluations, while
%we assume it should be more for EfficientNet-B0, on the other hand, 
\texttt{LLMatic} requires $2,000$ evaluations to find a competitive network. EfficientNet-B0 was first trained on the ImageNet dataset~\citep{deng2009imagenet} and then on CIFAR-10 via transfer learning~\citep{torrey2010transfer}. This is an advantage for EfficientNet-B0 as ImageNet has many classes and is an order of magnitude larger dataset.

Figure \ref{fig:archives} demonstrates how each archive is filled on average. We can see that the prompt archive contains high-performing individuals who have the first few prompts and higher temperatures. 
Some of the high-performing individuals do have lower temperatures, which suggests that sometimes it is useful to generate neural network layers in a less stochastic manner. For network archives, we observe a diversity of high-performing networks with respect to both FLOPS and width-to-depth ratio. More than $20$ individuals are competitive networks in this archive.

To investigate our choice of probabilities for crossover and mutation ($0.3$ and $0.7$, respectively), we observe the number of trainable networks generated per generation (see Figure \ref{fig:crossmut}). We use this as a measure, since the more functional individuals we have, the greater the chance of high-performing individuals. For this purpose, we train \texttt{LLMatic} with uniform probabilities, and $0.3$ for mutation and $0.7$ for crossover. We observe that uniform probabilities are still competitive with the original setting, while increasing the crossover probability makes it worse. The results of these experiments and results of the ablation study for \texttt{Crossover-Only-LLMatic} and \texttt{Mutation-Only-LLMatic} lead us to the conclusion that mutation should be given more probability of being selected.

\section{Experiments on NAS-bench-201}

% \newcommand{\rrr}[1]{0.32\linewidth}
% \begin{figure*}[!htb]
%     \centering
%     \begin{subfigure}{\rrr}
%         \includegraphics[width=1.01\textwidth]{Figures/cifar10acc.png}
%         \caption{CIFAR-10}
%         \label{c10}
%     \end{subfigure}
%     \begin{subfigure}{\rrr}
%         \includegraphics[width=1.01\textwidth]{Figures/cifar100acc.png}
%         \caption{CIFAR-100}
%         \label{c100}
%     \end{subfigure}
%     \begin{subfigure}{\rrr}
%         \includegraphics[width=1.01\textwidth]{Figures/imagenet100acc.png}
%         \caption{ImageNet16-120}
%         \label{im16}
%     \end{subfigure}
%     \caption{Illustration of test accuracies of all networks across all datasets and best-found networks in each generation by \texttt{LLMatic}.}
%     \label{fig:testacc}
% \end{figure*}

% \subsection{Dataset and Benchmark}
Next, we extend our experimentation of \texttt{LLMatic} to the NAS-bench-201 benchmark~\citep{dong2020bench}, which searches a cell block for a constant neural network structure. The structure is initiated with one $3 \times 3$ convolution with $16$ output channels and a batch normalization layer~\citep{ioffe2015batch}. The main body of the skeleton includes three stacks of cells, connected by a residual block. Each cell is stacked 5 times, with the number of output channels as $16$, $32$ and $64$ for the first, second and third stages, respectively. The intermediate residual block is the basic residual block with a stride of $2$~\citep{he2016deep}, which serves to downsample the spatial size and double the channels of an input feature map. The shortcut path in this residual block consists of a $2 \times 2$ average pooling layer with a stride of $2$ and a $1 \times 1$ convolution. The skeleton ends with a global average pooling layer to flatten the feature map into a feature vector. Classification uses a fully connected layer with a softmax layer to transform the feature vector into the final prediction.

The specified cell within the search domain is depicted as a densely connected directed acyclic graph with four nodes and six edges; here, nodes symbolise feature maps while edges denote operations. There are five possible operations: (1) zeroize, (2) skip connection, (3) $1 \times 1$ convolution, (4) $3 \times3$ convolution, and (5) $3 \times 3$ average pooling layer. Zeroize drops out the associated edge operation. Given five operations to choose from, the aggregate count of potential search spaces is $5^6 = 15625$ cell combinations. Evaluations are carried out on CIFAR10, CIFAR100~\citep{krizhevsky2009learning}, and ImageNet16-120~\citep{chrabaszcz2017downsampled}. ImageNet16-120 is a variant of ImageNet dataset~\citep{russakovsky2015imagenet} which is downsampled to 16x16 image sizes and contains the first 120 classes.

\begin{figure}[b!]
    \centering
    \begin{subfigure}{0.9\linewidth}
        \includegraphics[width=1.01\textwidth]{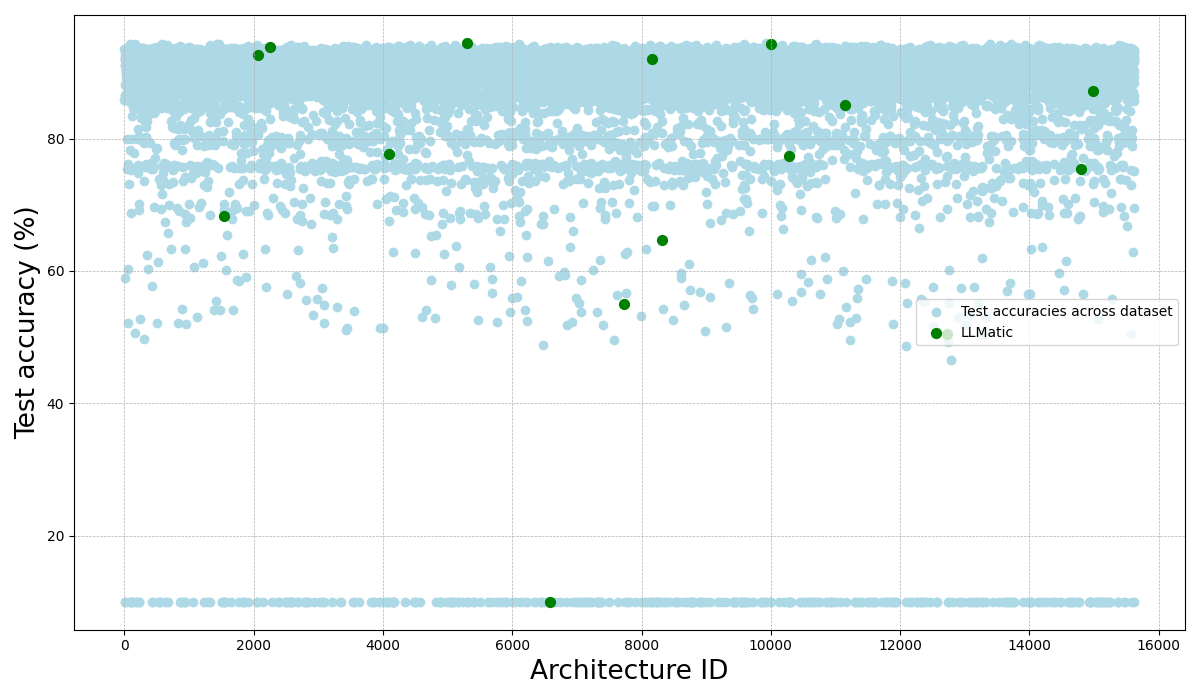}
        \caption{CIFAR-10}
        \label{c10}
    \end{subfigure}
    \\
    \begin{subfigure}{0.9\linewidth}
        \includegraphics[width=1.01\textwidth]{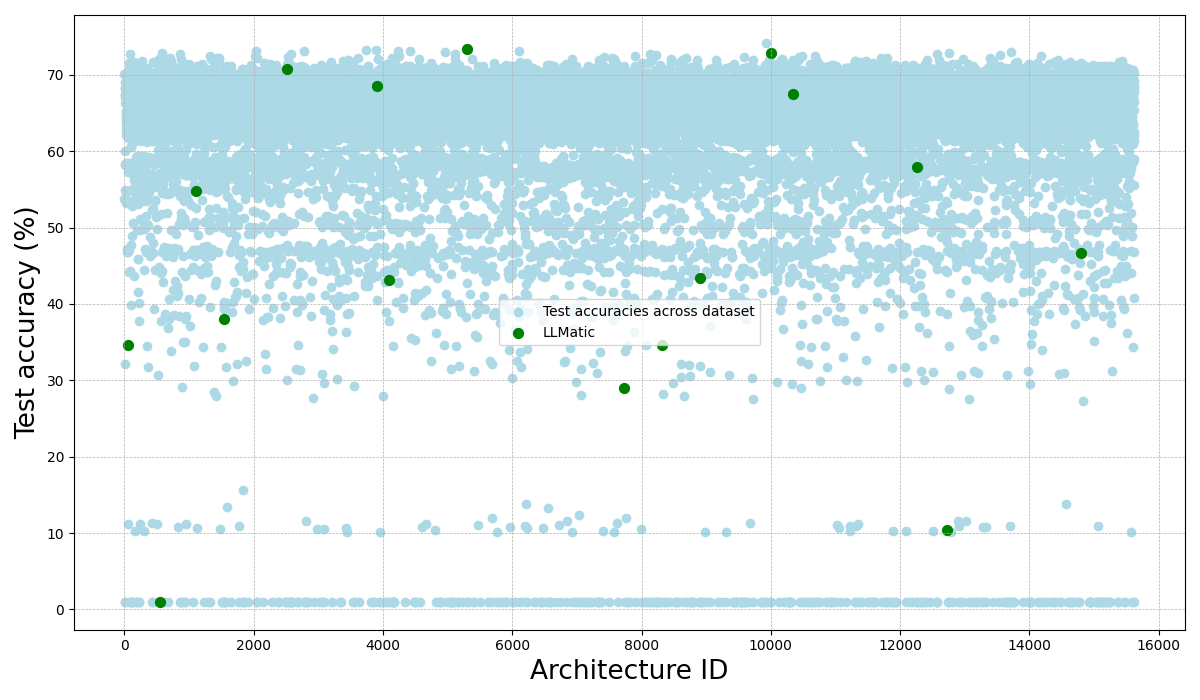}
        \caption{CIFAR-100}
        \label{c100}
    \end{subfigure}
    \\
    \begin{subfigure}{0.9\linewidth}
        \includegraphics[width=1.01\textwidth]{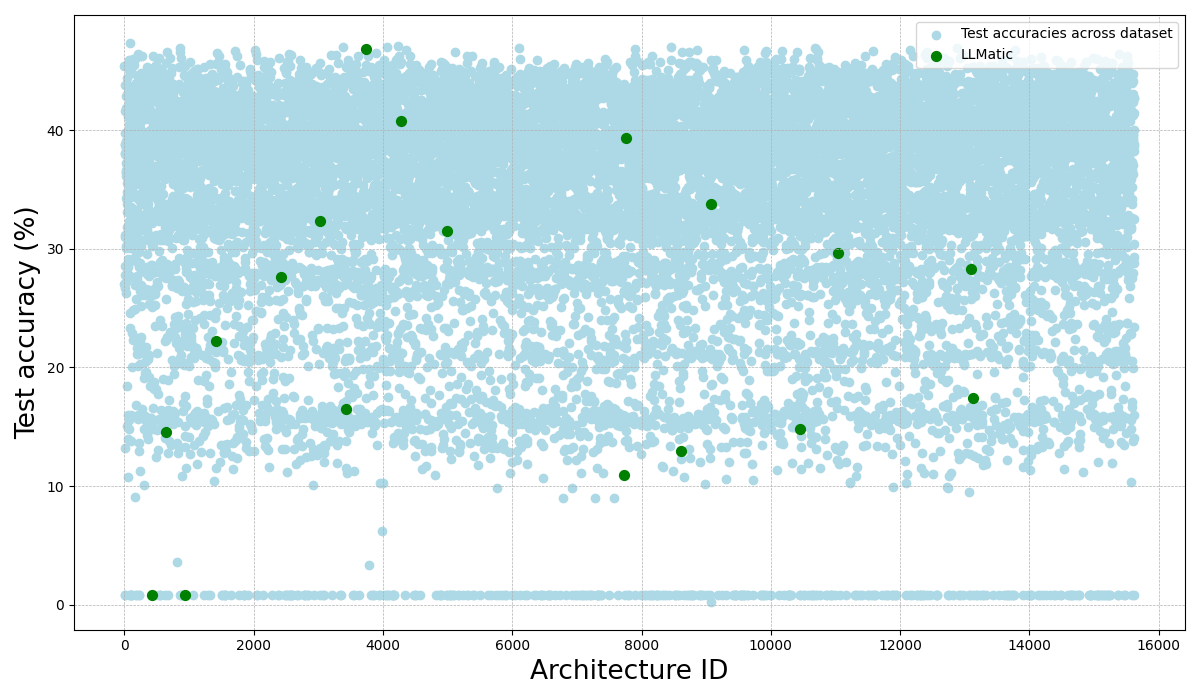}
        \caption{ImageNet16-120}
        \label{im16}
    \end{subfigure}
    \caption{Illustration of test accuracies of all networks across all datasets and best-found networks in each generation by \texttt{LLMatic}.}
    \label{fig:testacc}
\end{figure}

\subsection{Results}

To remain consistent with our previous experiments, \texttt{LLMatic} searches for $20$ generations and $100$ cells in a generation. We curate the prompt to cater for a controllable generation by restricting it to the five operations. Refer to supplementary material for an example of how we generate queryable cells. For our network archive, we take minimum and maximum FLOPS as the bounds for the behaviour descriptor.

\begin{table}[b!]
\caption{A comparison of test accuracy on the NAS-bench-201 benchmark. We provide the optimal accuracy for reference, which is the maximum accuracy that can be achieved in NAS-bench-201. The results for \texttt{LLMatic} are averaged over 10 runs.}
\label{nas201results}
\begin{center}
\begin{tabular}{llll}
\hline Method                               & CIFAR-10   & CIFAR-100  & ImageNet16-120 \\ \hline 
\multicolumn{1}{l|}{DARTS}           & 54.30±0.00 & 15.61±0.00 & 16.32±0.00     \\ 
\multicolumn{1}{l|}{Random Search}   & 93.70±0.36 & 71.04±1.07 & 44.57±1.25     \\ 
\multicolumn{1}{l|}{GENIUS}          & 93.79±0.09 & 70.91±0.72 & 44.96±1.02     \\  
\multicolumn{1}{l|}{$\Lambda$-DARTS} & 94.36±0.00 & 73.51±0.00 & 46.34±0.00     \\
\multicolumn{1}{l|}{LLMatic}         & 94.26±0.13 & 71.62±1.73 & 45.87±0.96 \\
\hline 
\multicolumn{1}{l|}{Optimal}         & 94.47 & 74.17 & 47.33 \\ \hline 
\end{tabular}
\end{center}
\end{table}

We compare our results with the GPT-4-based NAS algorithm GENIUS~\citep{zheng2023can}, which serves as an LLM baseline, as well as random search. We also compare to prior work, including DARTS ~\citep{liu2018darts} and $\Lambda$-DARTS~\citep{movahedi2022lambda}, which achieves near-optimal results. As \autoref{nas201results} indicates, \texttt{LLMatic} outperforms the GPT-4-based NAS, and produces results that are near state-of-the-art, although it has more variation than the competing methods.

Furthermore, in \autoref{fig:testacc} we investigate the networks discovered by \texttt{LLMatic} over each generation. We observe that the distribution of found networks is spread wide in the search space. This is due to the procedural nature and exploration capabilities of \texttt{LLMatic} through the prompt archive. To demonstrate near-to-optimal networks we illustrate in \autoref{tabrank} the maximum ranked networks based on test accuracies searched by \texttt{LLMatic}.

\begin{table}[]
\caption{Maximum rank achieved by \texttt{LLMatic} on each dataset in NAS-bench-201.}
\label{tabrank}
\begin{center}
\begin{tabular}{ll}
\hline
Method         & Rank \\ \hline
CIFAR-10       & 2    \\
CIFAR-100      & 2    \\
ImageNet16-120 & 11 \\
\hline
\end{tabular}
\end{center}
\end{table}

\section{Conclusion and Future Work}

To conclude, we present \texttt{LLMatic}: a novel neural architecture search (NAS) algorithm that harnesses the power of large language models (LLMs) and Quality-Diversity (QD) optimization algorithms. \texttt{LLMatic} successfully finds competitive networks that are diverse in architecture. We show empirically that \texttt{LLMatic} can find more than \textit{20} competitive networks in CIFAR-10 and near-to-optimal networks in NAS-bench-201, using only \textit{2000} evaluations. \texttt{LLMatic} decreases the max population size per generation to only \textit{100}. \texttt{LLMatic} achieves this while relying on a \textit{6.1B} parameter language model. Furthermore, we show that each component in \texttt{LLMatic} is necessary. We conducted an extensive ablation study and found that \texttt{LLMatic} finds the network with the best accuracy among other variants. 

\texttt{LLMatic} achieves this with many constraints in hand. Firstly, we use CodeGen-6.1B code generation LLM, which is a smaller language model when compared to existing LLMs. This demonstrates the computationally efficiency of \texttt{LLMatic}, and gives us reason to believe that further gains could be unlocked by incorporating larger language models
%and how much it can unlock the value with a larger language model.
Secondly, due to computational resources, we keep our searches to $2000$, and still find competitive networks. 

In future work, \texttt{LLMatic} should be compared to other NAS methods on other computer vision and natural language processing tasks. As neuroevolution is similar to NAS, \texttt{LLMatic} could be compared to reinforcement learning benchmarks as well. With this, \texttt{LLMatic} can be used in tasks such as open-ended learning as well \citep{nasir2022augmentative}.

\section{Acknowledgments}

The authors acknowledge the Centre for High Performance Computing (CHPC),
South Africa, for providing computational resources to this research project.

\bibliographystyle{ACM-Reference-Format}
\bibliography{main}


\begin{thebibliography}{53}

%%% the Web address.

\ifx \showCODEN    \undefined \def \showCODEN     #1{\unskip}     \fi
\ifx \showDOI      \undefined \def \showDOI       #1{#1}\fi
\ifx \showISBNx    \undefined \def \showISBNx     #1{\unskip}     \fi
\ifx \showISBNxiii \undefined \def \showISBNxiii  #1{\unskip}     \fi
\ifx \showISSN     \undefined \def \showISSN      #1{\unskip}     \fi
\ifx \showLCCN     \undefined \def \showLCCN      #1{\unskip}     \fi
\ifx \shownote     \undefined \def \shownote      #1{#1}          \fi
\ifx \showarticletitle \undefined \def \showarticletitle #1{#1}   \fi
\ifx \showURL      \undefined \def \showURL       {\relax}        \fi
% invisible to TeX
\providecommand\bibfield[2]{#2}
\providecommand\bibinfo[2]{#2}
\providecommand\natexlab[1]{#1}
\providecommand\showeprint[2][]{arXiv:#2}

\bibitem[\protect\citeauthoryear{Adelani, Alabi, Fan, Kreutzer, Shen, Reid, Ruiter, Klakow, Nabende, Chang, et~al\mbox{.}}{Adelani et~al\mbox{.}}{2022}]%
        {adelani2022few}
\bibfield{author}{\bibinfo{person}{David~Ifeoluwa Adelani}, \bibinfo{person}{Jesujoba~Oluwadara Alabi}, \bibinfo{person}{Angela Fan}, \bibinfo{person}{Julia Kreutzer}, \bibinfo{person}{Xiaoyu Shen}, \bibinfo{person}{Machel Reid}, \bibinfo{person}{Dana Ruiter}, \bibinfo{person}{Dietrich Klakow}, \bibinfo{person}{Peter Nabende}, \bibinfo{person}{Ernie Chang}, {et~al\mbox{.}}} \bibinfo{year}{2022}\natexlab{}.
\newblock \showarticletitle{A few thousand translations go a long way! leveraging pre-trained models for african news translation}.
\newblock \bibinfo{journal}{\emph{arXiv preprint arXiv:2205.02022}} (\bibinfo{year}{2022}).
\newblock


\bibitem[\protect\citeauthoryear{Ayala, Mu{\~n}oz, Llanos, and dos Santos~Coelho}{Ayala et~al\mbox{.}}{2017}]%
        {ayala2017efficient}
\bibfield{author}{\bibinfo{person}{Helon Vicente~Hultmann Ayala}, \bibinfo{person}{Daniel~M Mu{\~n}oz}, \bibinfo{person}{Carlos~H Llanos}, {and} \bibinfo{person}{Leandro dos Santos~Coelho}.} \bibinfo{year}{2017}\natexlab{}.
\newblock \showarticletitle{Efficient hardware implementation of radial basis function neural network with customized-precision floating-point operations}.
\newblock \bibinfo{journal}{\emph{Control Engineering Practice}}  \bibinfo{volume}{60} (\bibinfo{year}{2017}), \bibinfo{pages}{124--132}.
\newblock


\bibitem[\protect\citeauthoryear{Bentley}{Bentley}{1975}]%
        {bentley1975multidimensional}
\bibfield{author}{\bibinfo{person}{Jon~Louis Bentley}.} \bibinfo{year}{1975}\natexlab{}.
\newblock \showarticletitle{Multidimensional binary search trees used for associative searching}.
\newblock \bibinfo{journal}{\emph{Commun. ACM}} \bibinfo{volume}{18}, \bibinfo{number}{9} (\bibinfo{year}{1975}), \bibinfo{pages}{509--517}.
\newblock


\bibitem[\protect\citeauthoryear{Bottou}{Bottou}{2010}]%
        {bottou2010large}
\bibfield{author}{\bibinfo{person}{L{\'e}on Bottou}.} \bibinfo{year}{2010}\natexlab{}.
\newblock \showarticletitle{Large-scale machine learning with stochastic gradient descent}. In \bibinfo{booktitle}{\emph{Proceedings of COMPSTAT'2010: 19th International Conference on Computational StatisticsParis France, August 22-27, 2010 Keynote, Invited and Contributed Papers}}. Springer, \bibinfo{pages}{177--186}.
\newblock


\bibitem[\protect\citeauthoryear{Brown, Mann, Ryder, Subbiah, Kaplan, Dhariwal, Neelakantan, Shyam, Sastry, Askell, et~al\mbox{.}}{Brown et~al\mbox{.}}{2020}]%
        {brown2020language}
\bibfield{author}{\bibinfo{person}{Tom Brown}, \bibinfo{person}{Benjamin Mann}, \bibinfo{person}{Nick Ryder}, \bibinfo{person}{Melanie Subbiah}, \bibinfo{person}{Jared~D Kaplan}, \bibinfo{person}{Prafulla Dhariwal}, \bibinfo{person}{Arvind Neelakantan}, \bibinfo{person}{Pranav Shyam}, \bibinfo{person}{Girish Sastry}, \bibinfo{person}{Amanda Askell}, {et~al\mbox{.}}} \bibinfo{year}{2020}\natexlab{}.
\newblock \showarticletitle{Language models are few-shot learners}.
\newblock \bibinfo{journal}{\emph{Advances in neural information processing systems}}  \bibinfo{volume}{33} (\bibinfo{year}{2020}), \bibinfo{pages}{1877--1901}.
\newblock


\bibitem[\protect\citeauthoryear{Chen, Dohan, and So}{Chen et~al\mbox{.}}{2023}]%
        {chen2023evoprompting}
\bibfield{author}{\bibinfo{person}{Angelica Chen}, \bibinfo{person}{David~M Dohan}, {and} \bibinfo{person}{David~R So}.} \bibinfo{year}{2023}\natexlab{}.
\newblock \showarticletitle{EvoPrompting: Language Models for Code-Level Neural Architecture Search}.
\newblock \bibinfo{journal}{\emph{arXiv preprint arXiv:2302.14838}} (\bibinfo{year}{2023}).
\newblock


\bibitem[\protect\citeauthoryear{Chen, Tworek, Jun, Yuan, Pinto, Kaplan, Edwards, Burda, Joseph, Brockman, et~al\mbox{.}}{Chen et~al\mbox{.}}{2021}]%
        {chen2021evaluating}
\bibfield{author}{\bibinfo{person}{Mark Chen}, \bibinfo{person}{Jerry Tworek}, \bibinfo{person}{Heewoo Jun}, \bibinfo{person}{Qiming Yuan}, \bibinfo{person}{Henrique Ponde de~Oliveira Pinto}, \bibinfo{person}{Jared Kaplan}, \bibinfo{person}{Harri Edwards}, \bibinfo{person}{Yuri Burda}, \bibinfo{person}{Nicholas Joseph}, \bibinfo{person}{Greg Brockman}, {et~al\mbox{.}}} \bibinfo{year}{2021}\natexlab{}.
\newblock \showarticletitle{Evaluating large language models trained on code}.
\newblock \bibinfo{journal}{\emph{arXiv preprint arXiv:2107.03374}} (\bibinfo{year}{2021}).
\newblock


\bibitem[\protect\citeauthoryear{Chrabaszcz, Loshchilov, and Hutter}{Chrabaszcz et~al\mbox{.}}{2017}]%
        {chrabaszcz2017downsampled}
\bibfield{author}{\bibinfo{person}{Patryk Chrabaszcz}, \bibinfo{person}{Ilya Loshchilov}, {and} \bibinfo{person}{Frank Hutter}.} \bibinfo{year}{2017}\natexlab{}.
\newblock \showarticletitle{A downsampled variant of imagenet as an alternative to the cifar datasets}.
\newblock \bibinfo{journal}{\emph{arXiv preprint arXiv:1707.08819}} (\bibinfo{year}{2017}).
\newblock


\bibitem[\protect\citeauthoryear{Cully, Clune, Tarapore, and Mouret}{Cully et~al\mbox{.}}{2015}]%
        {cully2015robots}
\bibfield{author}{\bibinfo{person}{Antoine Cully}, \bibinfo{person}{Jeff Clune}, \bibinfo{person}{Danesh Tarapore}, {and} \bibinfo{person}{Jean-Baptiste Mouret}.} \bibinfo{year}{2015}\natexlab{}.
\newblock \showarticletitle{Robots that can adapt like animals}.
\newblock \bibinfo{journal}{\emph{Nature}} \bibinfo{volume}{521}, \bibinfo{number}{7553} (\bibinfo{year}{2015}), \bibinfo{pages}{503--507}.
\newblock


\bibitem[\protect\citeauthoryear{Cully and Demiris}{Cully and Demiris}{2017}]%
        {cully2017quality}
\bibfield{author}{\bibinfo{person}{Antoine Cully} {and} \bibinfo{person}{Yiannis Demiris}.} \bibinfo{year}{2017}\natexlab{}.
\newblock \showarticletitle{Quality and diversity optimization: A unifying modular framework}.
\newblock \bibinfo{journal}{\emph{IEEE Transactions on Evolutionary Computation}} \bibinfo{volume}{22}, \bibinfo{number}{2} (\bibinfo{year}{2017}), \bibinfo{pages}{245--259}.
\newblock


\bibitem[\protect\citeauthoryear{Deng, Dong, Socher, Li, Li, and Fei-Fei}{Deng et~al\mbox{.}}{2009}]%
        {deng2009imagenet}
\bibfield{author}{\bibinfo{person}{Jia Deng}, \bibinfo{person}{Wei Dong}, \bibinfo{person}{Richard Socher}, \bibinfo{person}{Li-Jia Li}, \bibinfo{person}{Kai Li}, {and} \bibinfo{person}{Li Fei-Fei}.} \bibinfo{year}{2009}\natexlab{}.
\newblock \showarticletitle{Imagenet: A large-scale hierarchical image database}. In \bibinfo{booktitle}{\emph{2009 IEEE conference on computer vision and pattern recognition}}. Ieee, \bibinfo{pages}{248--255}.
\newblock


\bibitem[\protect\citeauthoryear{Dokmanic, Parhizkar, Ranieri, and Vetterli}{Dokmanic et~al\mbox{.}}{2015}]%
        {dokmanic2015euclidean}
\bibfield{author}{\bibinfo{person}{Ivan Dokmanic}, \bibinfo{person}{Reza Parhizkar}, \bibinfo{person}{Juri Ranieri}, {and} \bibinfo{person}{Martin Vetterli}.} \bibinfo{year}{2015}\natexlab{}.
\newblock \showarticletitle{Euclidean distance matrices: essential theory, algorithms, and applications}.
\newblock \bibinfo{journal}{\emph{IEEE Signal Processing Magazine}} \bibinfo{volume}{32}, \bibinfo{number}{6} (\bibinfo{year}{2015}), \bibinfo{pages}{12--30}.
\newblock


\bibitem[\protect\citeauthoryear{Dong and Yang}{Dong and Yang}{2020}]%
        {dong2020bench}
\bibfield{author}{\bibinfo{person}{Xuanyi Dong} {and} \bibinfo{person}{Yi Yang}.} \bibinfo{year}{2020}\natexlab{}.
\newblock \showarticletitle{Nas-bench-201: Extending the scope of reproducible neural architecture search}.
\newblock \bibinfo{journal}{\emph{arXiv preprint arXiv:2001.00326}} (\bibinfo{year}{2020}).
\newblock


\bibitem[\protect\citeauthoryear{Elsken, Metzen, and Hutter}{Elsken et~al\mbox{.}}{2019}]%
        {elsken2019neural}
\bibfield{author}{\bibinfo{person}{Thomas Elsken}, \bibinfo{person}{Jan~Hendrik Metzen}, {and} \bibinfo{person}{Frank Hutter}.} \bibinfo{year}{2019}\natexlab{}.
\newblock \showarticletitle{Neural architecture search: A survey}.
\newblock \bibinfo{journal}{\emph{The Journal of Machine Learning Research}} \bibinfo{volume}{20}, \bibinfo{number}{1} (\bibinfo{year}{2019}), \bibinfo{pages}{1997--2017}.
\newblock


\bibitem[\protect\citeauthoryear{Gaier, Asteroth, and Mouret}{Gaier et~al\mbox{.}}{2019}]%
        {gaier2019quality}
\bibfield{author}{\bibinfo{person}{Adam Gaier}, \bibinfo{person}{Alexander Asteroth}, {and} \bibinfo{person}{Jean-Baptiste Mouret}.} \bibinfo{year}{2019}\natexlab{}.
\newblock \showarticletitle{Are quality diversity algorithms better at generating stepping stones than objective-based search?}. In \bibinfo{booktitle}{\emph{Proceedings of the Genetic and Evolutionary Computation Conference Companion}}. \bibinfo{pages}{115--116}.
\newblock


\bibitem[\protect\citeauthoryear{Gao, Biderman, Black, Golding, Hoppe, Foster, Phang, He, Thite, Nabeshima, et~al\mbox{.}}{Gao et~al\mbox{.}}{2020}]%
        {gao2020pile}
\bibfield{author}{\bibinfo{person}{Leo Gao}, \bibinfo{person}{Stella Biderman}, \bibinfo{person}{Sid Black}, \bibinfo{person}{Laurence Golding}, \bibinfo{person}{Travis Hoppe}, \bibinfo{person}{Charles Foster}, \bibinfo{person}{Jason Phang}, \bibinfo{person}{Horace He}, \bibinfo{person}{Anish Thite}, \bibinfo{person}{Noa Nabeshima}, {et~al\mbox{.}}} \bibinfo{year}{2020}\natexlab{}.
\newblock \showarticletitle{The pile: An 800gb dataset of diverse text for language modeling}.
\newblock \bibinfo{journal}{\emph{arXiv preprint arXiv:2101.00027}} (\bibinfo{year}{2020}).
\newblock


\bibitem[\protect\citeauthoryear{Greydanus}{Greydanus}{2020}]%
        {greydanus2020scaling}
\bibfield{author}{\bibinfo{person}{Sam Greydanus}.} \bibinfo{year}{2020}\natexlab{}.
\newblock \showarticletitle{Scaling down deep learning}.
\newblock \bibinfo{journal}{\emph{arXiv preprint arXiv:2011.14439}} (\bibinfo{year}{2020}).
\newblock


\bibitem[\protect\citeauthoryear{He, Zhang, Ren, and Sun}{He et~al\mbox{.}}{2016}]%
        {he2016deep}
\bibfield{author}{\bibinfo{person}{Kaiming He}, \bibinfo{person}{Xiangyu Zhang}, \bibinfo{person}{Shaoqing Ren}, {and} \bibinfo{person}{Jian Sun}.} \bibinfo{year}{2016}\natexlab{}.
\newblock \showarticletitle{Deep residual learning for image recognition}. In \bibinfo{booktitle}{\emph{Proceedings of the IEEE conference on computer vision and pattern recognition}}. \bibinfo{pages}{770--778}.
\newblock


\bibitem[\protect\citeauthoryear{Ioffe and Szegedy}{Ioffe and Szegedy}{2015}]%
        {ioffe2015batch}
\bibfield{author}{\bibinfo{person}{Sergey Ioffe} {and} \bibinfo{person}{Christian Szegedy}.} \bibinfo{year}{2015}\natexlab{}.
\newblock \showarticletitle{Batch normalization: Accelerating deep network training by reducing internal covariate shift}. In \bibinfo{booktitle}{\emph{International conference on machine learning}}. pmlr, \bibinfo{pages}{448--456}.
\newblock


\bibitem[\protect\citeauthoryear{Jaafra, Laurent, Deruyver, and Naceur}{Jaafra et~al\mbox{.}}{2019}]%
        {jaafra2019reinforcement}
\bibfield{author}{\bibinfo{person}{Yesmina Jaafra}, \bibinfo{person}{Jean~Luc Laurent}, \bibinfo{person}{Aline Deruyver}, {and} \bibinfo{person}{Mohamed~Saber Naceur}.} \bibinfo{year}{2019}\natexlab{}.
\newblock \showarticletitle{Reinforcement learning for neural architecture search: A review}.
\newblock \bibinfo{journal}{\emph{Image and Vision Computing}}  \bibinfo{volume}{89} (\bibinfo{year}{2019}), \bibinfo{pages}{57--66}.
\newblock


\bibitem[\protect\citeauthoryear{Kandasamy, Neiswanger, Schneider, Poczos, and Xing}{Kandasamy et~al\mbox{.}}{2018}]%
        {kandasamy2018neural}
\bibfield{author}{\bibinfo{person}{Kirthevasan Kandasamy}, \bibinfo{person}{Willie Neiswanger}, \bibinfo{person}{Jeff Schneider}, \bibinfo{person}{Barnabas Poczos}, {and} \bibinfo{person}{Eric~P Xing}.} \bibinfo{year}{2018}\natexlab{}.
\newblock \showarticletitle{Neural architecture search with bayesian optimisation and optimal transport}.
\newblock \bibinfo{journal}{\emph{Advances in neural information processing systems}}  \bibinfo{volume}{31} (\bibinfo{year}{2018}).
\newblock


\bibitem[\protect\citeauthoryear{Krizhevsky, Hinton, et~al\mbox{.}}{Krizhevsky et~al\mbox{.}}{2009}]%
        {krizhevsky2009learning}
\bibfield{author}{\bibinfo{person}{Alex Krizhevsky}, \bibinfo{person}{Geoffrey Hinton}, {et~al\mbox{.}}} \bibinfo{year}{2009}\natexlab{}.
\newblock \showarticletitle{Learning multiple layers of features from tiny images}.
\newblock  (\bibinfo{year}{2009}).
\newblock


\bibitem[\protect\citeauthoryear{Lehman, Gordon, Jain, Ndousse, Yeh, and Stanley}{Lehman et~al\mbox{.}}{2022}]%
        {lehman2022evolution}
\bibfield{author}{\bibinfo{person}{Joel Lehman}, \bibinfo{person}{Jonathan Gordon}, \bibinfo{person}{Shawn Jain}, \bibinfo{person}{Kamal Ndousse}, \bibinfo{person}{Cathy Yeh}, {and} \bibinfo{person}{Kenneth~O Stanley}.} \bibinfo{year}{2022}\natexlab{}.
\newblock \showarticletitle{Evolution through large models}.
\newblock \bibinfo{journal}{\emph{arXiv preprint arXiv:2206.08896}} (\bibinfo{year}{2022}).
\newblock


\bibitem[\protect\citeauthoryear{Liu, Simonyan, and Yang}{Liu et~al\mbox{.}}{2018}]%
        {liu2018darts}
\bibfield{author}{\bibinfo{person}{Hanxiao Liu}, \bibinfo{person}{Karen Simonyan}, {and} \bibinfo{person}{Yiming Yang}.} \bibinfo{year}{2018}\natexlab{}.
\newblock \showarticletitle{Darts: Differentiable architecture search}.
\newblock \bibinfo{journal}{\emph{arXiv preprint arXiv:1806.09055}} (\bibinfo{year}{2018}).
\newblock


\bibitem[\protect\citeauthoryear{Liu, Sun, Xue, Zhang, Yen, and Tan}{Liu et~al\mbox{.}}{2021}]%
        {liu2021survey}
\bibfield{author}{\bibinfo{person}{Yuqiao Liu}, \bibinfo{person}{Yanan Sun}, \bibinfo{person}{Bing Xue}, \bibinfo{person}{Mengjie Zhang}, \bibinfo{person}{Gary~G Yen}, {and} \bibinfo{person}{Kay~Chen Tan}.} \bibinfo{year}{2021}\natexlab{}.
\newblock \showarticletitle{A survey on evolutionary neural architecture search}.
\newblock \bibinfo{journal}{\emph{IEEE transactions on neural networks and learning systems}} (\bibinfo{year}{2021}).
\newblock


\bibitem[\protect\citeauthoryear{Mehta, White, Zela, Krishnakumar, Zabergja, Moradian, Safari, Yu, and Hutter}{Mehta et~al\mbox{.}}{2022}]%
        {mehta2022bench}
\bibfield{author}{\bibinfo{person}{Yash Mehta}, \bibinfo{person}{Colin White}, \bibinfo{person}{Arber Zela}, \bibinfo{person}{Arjun Krishnakumar}, \bibinfo{person}{Guri Zabergja}, \bibinfo{person}{Shakiba Moradian}, \bibinfo{person}{Mahmoud Safari}, \bibinfo{person}{Kaicheng Yu}, {and} \bibinfo{person}{Frank Hutter}.} \bibinfo{year}{2022}\natexlab{}.
\newblock \showarticletitle{NAS-Bench-Suite: NAS evaluation is (now) surprisingly easy}.
\newblock \bibinfo{journal}{\emph{arXiv preprint arXiv:2201.13396}} (\bibinfo{year}{2022}).
\newblock


\bibitem[\protect\citeauthoryear{Miller, Todd, and Hegde}{Miller et~al\mbox{.}}{1989}]%
        {miller1989designing}
\bibfield{author}{\bibinfo{person}{Geoffrey~F Miller}, \bibinfo{person}{Peter~M Todd}, {and} \bibinfo{person}{Shailesh~U Hegde}.} \bibinfo{year}{1989}\natexlab{}.
\newblock \showarticletitle{Designing Neural Networks Using Genetic Algorithms.}. In \bibinfo{booktitle}{\emph{ICGA}}, Vol.~\bibinfo{volume}{89}. \bibinfo{pages}{379--384}.
\newblock


\bibitem[\protect\citeauthoryear{Mouret and Clune}{Mouret and Clune}{2015}]%
        {mouret2015illuminating}
\bibfield{author}{\bibinfo{person}{Jean-Baptiste Mouret} {and} \bibinfo{person}{Jeff Clune}.} \bibinfo{year}{2015}\natexlab{}.
\newblock \showarticletitle{Illuminating search spaces by mapping elites}.
\newblock \bibinfo{journal}{\emph{arXiv preprint arXiv:1504.04909}} (\bibinfo{year}{2015}).
\newblock


\bibitem[\protect\citeauthoryear{Movahedi, Adabinejad, Imani, Keshavarz, Dehghani, Shakery, and Araabi}{Movahedi et~al\mbox{.}}{2022}]%
        {movahedi2022lambda}
\bibfield{author}{\bibinfo{person}{Sajad Movahedi}, \bibinfo{person}{Melika Adabinejad}, \bibinfo{person}{Ayyoob Imani}, \bibinfo{person}{Arezou Keshavarz}, \bibinfo{person}{Mostafa Dehghani}, \bibinfo{person}{Azadeh Shakery}, {and} \bibinfo{person}{Babak~N Araabi}.} \bibinfo{year}{2022}\natexlab{}.
\newblock \showarticletitle{{$\Delta$}-DARTS: Mitigating Performance Collapse by Harmonizing Operation Selection among Cells}.
\newblock \bibinfo{journal}{\emph{arXiv preprint arXiv:2210.07998}} (\bibinfo{year}{2022}).
\newblock


\bibitem[\protect\citeauthoryear{Nair and Hinton}{Nair and Hinton}{2010}]%
        {nair2010rectified}
\bibfield{author}{\bibinfo{person}{Vinod Nair} {and} \bibinfo{person}{Geoffrey~E Hinton}.} \bibinfo{year}{2010}\natexlab{}.
\newblock \showarticletitle{Rectified linear units improve restricted boltzmann machines}. In \bibinfo{booktitle}{\emph{Proceedings of the 27th international conference on machine learning (ICML-10)}}. \bibinfo{pages}{807--814}.
\newblock


\bibitem[\protect\citeauthoryear{Nasir, Beukman, James, and Cleghorn}{Nasir et~al\mbox{.}}{2022}]%
        {nasir2022augmentative}
\bibfield{author}{\bibinfo{person}{Muhammad~Umair Nasir}, \bibinfo{person}{Michael Beukman}, \bibinfo{person}{Steven James}, {and} \bibinfo{person}{Christopher~Wesley Cleghorn}.} \bibinfo{year}{2022}\natexlab{}.
\newblock \showarticletitle{Augmentative Topology Agents For Open-ended Learning}.
\newblock \bibinfo{journal}{\emph{arXiv preprint arXiv:2210.11442}} (\bibinfo{year}{2022}).
\newblock


\bibitem[\protect\citeauthoryear{Nasir and Mchechesi}{Nasir and Mchechesi}{2022}]%
        {nasir2022geographical}
\bibfield{author}{\bibinfo{person}{Muhammad~Umair Nasir} {and} \bibinfo{person}{Innocent~Amos Mchechesi}.} \bibinfo{year}{2022}\natexlab{}.
\newblock \showarticletitle{Geographical distance is the new hyperparameter: A case study of finding the optimal pre-trained language for English-isiZulu machine translation}.
\newblock \bibinfo{journal}{\emph{arXiv preprint arXiv:2205.08621}} (\bibinfo{year}{2022}).
\newblock


\bibitem[\protect\citeauthoryear{Nasir and Togelius}{Nasir and Togelius}{2023}]%
        {nasir2023practical}
\bibfield{author}{\bibinfo{person}{Muhammad~U Nasir} {and} \bibinfo{person}{Julian Togelius}.} \bibinfo{year}{2023}\natexlab{}.
\newblock \showarticletitle{Practical PCG Through Large Language Models}.
\newblock \bibinfo{journal}{\emph{arXiv preprint arXiv:2305.18243}} (\bibinfo{year}{2023}).
\newblock


\bibitem[\protect\citeauthoryear{Nijkamp, Pang, Hayashi, Tu, Wang, Zhou, Savarese, and Xiong}{Nijkamp et~al\mbox{.}}{2022}]%
        {nijkamp2022codegen}
\bibfield{author}{\bibinfo{person}{Erik Nijkamp}, \bibinfo{person}{Bo Pang}, \bibinfo{person}{Hiroaki Hayashi}, \bibinfo{person}{Lifu Tu}, \bibinfo{person}{Huan Wang}, \bibinfo{person}{Yingbo Zhou}, \bibinfo{person}{Silvio Savarese}, {and} \bibinfo{person}{Caiming Xiong}.} \bibinfo{year}{2022}\natexlab{}.
\newblock \showarticletitle{Codegen: An open large language model for code with multi-turn program synthesis}.
\newblock \bibinfo{journal}{\emph{arXiv preprint arXiv:2203.13474}} (\bibinfo{year}{2022}).
\newblock


\bibitem[\protect\citeauthoryear{Paszke, Gross, Massa, Lerer, Bradbury, Chanan, Killeen, Lin, Gimelshein, Antiga, et~al\mbox{.}}{Paszke et~al\mbox{.}}{2019}]%
        {paszke2019pytorch}
\bibfield{author}{\bibinfo{person}{Adam Paszke}, \bibinfo{person}{Sam Gross}, \bibinfo{person}{Francisco Massa}, \bibinfo{person}{Adam Lerer}, \bibinfo{person}{James Bradbury}, \bibinfo{person}{Gregory Chanan}, \bibinfo{person}{Trevor Killeen}, \bibinfo{person}{Zeming Lin}, \bibinfo{person}{Natalia Gimelshein}, \bibinfo{person}{Luca Antiga}, {et~al\mbox{.}}} \bibinfo{year}{2019}\natexlab{}.
\newblock \showarticletitle{Pytorch: An imperative style, high-performance deep learning library}.
\newblock \bibinfo{journal}{\emph{Advances in neural information processing systems}}  \bibinfo{volume}{32} (\bibinfo{year}{2019}).
\newblock


\bibitem[\protect\citeauthoryear{Pugh, Soros, and Stanley}{Pugh et~al\mbox{.}}{2016}]%
        {pugh2016quality}
\bibfield{author}{\bibinfo{person}{Justin~K Pugh}, \bibinfo{person}{Lisa~B Soros}, {and} \bibinfo{person}{Kenneth~O Stanley}.} \bibinfo{year}{2016}\natexlab{}.
\newblock \showarticletitle{Quality diversity: A new frontier for evolutionary computation}.
\newblock \bibinfo{journal}{\emph{Frontiers in Robotics and AI}} (\bibinfo{year}{2016}), \bibinfo{pages}{40}.
\newblock


\bibitem[\protect\citeauthoryear{Radford, Wu, Child, Luan, Amodei, Sutskever, et~al\mbox{.}}{Radford et~al\mbox{.}}{2019}]%
        {radford2019language}
\bibfield{author}{\bibinfo{person}{Alec Radford}, \bibinfo{person}{Jeffrey Wu}, \bibinfo{person}{Rewon Child}, \bibinfo{person}{David Luan}, \bibinfo{person}{Dario Amodei}, \bibinfo{person}{Ilya Sutskever}, {et~al\mbox{.}}} \bibinfo{year}{2019}\natexlab{}.
\newblock \showarticletitle{Language models are unsupervised multitask learners}.
\newblock \bibinfo{journal}{\emph{OpenAI blog}} \bibinfo{volume}{1}, \bibinfo{number}{8} (\bibinfo{year}{2019}), \bibinfo{pages}{9}.
\newblock


\bibitem[\protect\citeauthoryear{Russakovsky, Deng, Su, Krause, Satheesh, Ma, Huang, Karpathy, Khosla, Bernstein, et~al\mbox{.}}{Russakovsky et~al\mbox{.}}{2015}]%
        {russakovsky2015imagenet}
\bibfield{author}{\bibinfo{person}{Olga Russakovsky}, \bibinfo{person}{Jia Deng}, \bibinfo{person}{Hao Su}, \bibinfo{person}{Jonathan Krause}, \bibinfo{person}{Sanjeev Satheesh}, \bibinfo{person}{Sean Ma}, \bibinfo{person}{Zhiheng Huang}, \bibinfo{person}{Andrej Karpathy}, \bibinfo{person}{Aditya Khosla}, \bibinfo{person}{Michael Bernstein}, {et~al\mbox{.}}} \bibinfo{year}{2015}\natexlab{}.
\newblock \showarticletitle{Imagenet large scale visual recognition challenge}.
\newblock \bibinfo{journal}{\emph{International journal of computer vision}}  \bibinfo{volume}{115} (\bibinfo{year}{2015}), \bibinfo{pages}{211--252}.
\newblock


\bibitem[\protect\citeauthoryear{Shinn, Cassano, Gopinath, Narasimhan, and Yao}{Shinn et~al\mbox{.}}{2023}]%
        {shinn2023reflexion}
\bibfield{author}{\bibinfo{person}{Noah Shinn}, \bibinfo{person}{Federico Cassano}, \bibinfo{person}{Ashwin Gopinath}, \bibinfo{person}{Karthik~R Narasimhan}, {and} \bibinfo{person}{Shunyu Yao}.} \bibinfo{year}{2023}\natexlab{}.
\newblock \showarticletitle{Reflexion: Language agents with verbal reinforcement learning}. In \bibinfo{booktitle}{\emph{Thirty-seventh Conference on Neural Information Processing Systems}}.
\newblock


\bibitem[\protect\citeauthoryear{Stanley and Miikkulainen}{Stanley and Miikkulainen}{2002}]%
        {stanley2002evolving}
\bibfield{author}{\bibinfo{person}{Kenneth~O Stanley} {and} \bibinfo{person}{Risto Miikkulainen}.} \bibinfo{year}{2002}\natexlab{}.
\newblock \showarticletitle{Evolving neural networks through augmenting topologies}.
\newblock \bibinfo{journal}{\emph{Evolutionary computation}} \bibinfo{volume}{10}, \bibinfo{number}{2} (\bibinfo{year}{2002}), \bibinfo{pages}{99--127}.
\newblock


\bibitem[\protect\citeauthoryear{Tan, Chen, Pang, Vasudevan, Sandler, Howard, and Le}{Tan et~al\mbox{.}}{2019}]%
        {tan2019mnasnet}
\bibfield{author}{\bibinfo{person}{Mingxing Tan}, \bibinfo{person}{Bo Chen}, \bibinfo{person}{Ruoming Pang}, \bibinfo{person}{Vijay Vasudevan}, \bibinfo{person}{Mark Sandler}, \bibinfo{person}{Andrew Howard}, {and} \bibinfo{person}{Quoc~V Le}.} \bibinfo{year}{2019}\natexlab{}.
\newblock \showarticletitle{Mnasnet: Platform-aware neural architecture search for mobile}. In \bibinfo{booktitle}{\emph{Proceedings of the IEEE/CVF conference on computer vision and pattern recognition}}. \bibinfo{pages}{2820--2828}.
\newblock


\bibitem[\protect\citeauthoryear{Tan and Le}{Tan and Le}{2019}]%
        {tan2019efficientnet}
\bibfield{author}{\bibinfo{person}{Mingxing Tan} {and} \bibinfo{person}{Quoc Le}.} \bibinfo{year}{2019}\natexlab{}.
\newblock \showarticletitle{Efficientnet: Rethinking model scaling for convolutional neural networks}. In \bibinfo{booktitle}{\emph{International conference on machine learning}}. PMLR, \bibinfo{pages}{6105--6114}.
\newblock


\bibitem[\protect\citeauthoryear{Tenorio and Lee}{Tenorio and Lee}{1988}]%
        {tenorio1988self}
\bibfield{author}{\bibinfo{person}{Manoel Tenorio} {and} \bibinfo{person}{Wei-Tsih Lee}.} \bibinfo{year}{1988}\natexlab{}.
\newblock \showarticletitle{Self organizing neural networks for the identification problem}.
\newblock \bibinfo{journal}{\emph{Advances in Neural Information Processing Systems}}  \bibinfo{volume}{1} (\bibinfo{year}{1988}).
\newblock


\bibitem[\protect\citeauthoryear{Todd, Earle, Nasir, Green, and Togelius}{Todd et~al\mbox{.}}{2023}]%
        {todd2023level}
\bibfield{author}{\bibinfo{person}{Graham Todd}, \bibinfo{person}{Sam Earle}, \bibinfo{person}{Muhammad~Umair Nasir}, \bibinfo{person}{Michael~Cerny Green}, {and} \bibinfo{person}{Julian Togelius}.} \bibinfo{year}{2023}\natexlab{}.
\newblock \showarticletitle{Level Generation Through Large Language Models}. In \bibinfo{booktitle}{\emph{Proceedings of the 18th International Conference on the Foundations of Digital Games}}. \bibinfo{pages}{1--8}.
\newblock


\bibitem[\protect\citeauthoryear{Torrey and Shavlik}{Torrey and Shavlik}{2010}]%
        {torrey2010transfer}
\bibfield{author}{\bibinfo{person}{Lisa Torrey} {and} \bibinfo{person}{Jude Shavlik}.} \bibinfo{year}{2010}\natexlab{}.
\newblock \showarticletitle{Transfer learning}.
\newblock In \bibinfo{booktitle}{\emph{Handbook of research on machine learning applications and trends: algorithms, methods, and techniques}}. \bibinfo{publisher}{IGI global}, \bibinfo{pages}{242--264}.
\newblock


\bibitem[\protect\citeauthoryear{Ullah, Salam, and Masood}{Ullah et~al\mbox{.}}{2022}]%
        {ullah2022analysis}
\bibfield{author}{\bibinfo{person}{Sami Ullah}, \bibinfo{person}{Abdus Salam}, {and} \bibinfo{person}{Mohsin Masood}.} \bibinfo{year}{2022}\natexlab{}.
\newblock \showarticletitle{Analysis and comparison of a proposed mutation operator and its effects on the performance of genetic algorithm}.
\newblock \bibinfo{journal}{\emph{Indonesian Journal of Electrical Engineering and Computer Science}} \bibinfo{volume}{25}, \bibinfo{number}{2} (\bibinfo{year}{2022}), \bibinfo{pages}{1208--12168}.
\newblock


\bibitem[\protect\citeauthoryear{Vassiliades, Chatzilygeroudis, and Mouret}{Vassiliades et~al\mbox{.}}{2017}]%
        {vassiliades2017using}
\bibfield{author}{\bibinfo{person}{Vassilis Vassiliades}, \bibinfo{person}{Konstantinos Chatzilygeroudis}, {and} \bibinfo{person}{Jean-Baptiste Mouret}.} \bibinfo{year}{2017}\natexlab{}.
\newblock \showarticletitle{Using centroidal voronoi tessellations to scale up the multidimensional archive of phenotypic elites algorithm}.
\newblock \bibinfo{journal}{\emph{IEEE Transactions on Evolutionary Computation}} \bibinfo{volume}{22}, \bibinfo{number}{4} (\bibinfo{year}{2017}), \bibinfo{pages}{623--630}.
\newblock


\bibitem[\protect\citeauthoryear{Vaswani, Shazeer, Parmar, Uszkoreit, Jones, Gomez, Kaiser, and Polosukhin}{Vaswani et~al\mbox{.}}{2017}]%
        {vaswani2017attention}
\bibfield{author}{\bibinfo{person}{Ashish Vaswani}, \bibinfo{person}{Noam Shazeer}, \bibinfo{person}{Niki Parmar}, \bibinfo{person}{Jakob Uszkoreit}, \bibinfo{person}{Llion Jones}, \bibinfo{person}{Aidan~N Gomez}, \bibinfo{person}{{\L}ukasz Kaiser}, {and} \bibinfo{person}{Illia Polosukhin}.} \bibinfo{year}{2017}\natexlab{}.
\newblock \showarticletitle{Attention is all you need}.
\newblock \bibinfo{journal}{\emph{Advances in neural information processing systems}}  \bibinfo{volume}{30} (\bibinfo{year}{2017}).
\newblock


\bibitem[\protect\citeauthoryear{Veli{\v{c}}kovi{\'c}, Badia, Budden, Pascanu, Banino, Dashevskiy, Hadsell, and Blundell}{Veli{\v{c}}kovi{\'c} et~al\mbox{.}}{2022}]%
        {velivckovic2022clrs}
\bibfield{author}{\bibinfo{person}{Petar Veli{\v{c}}kovi{\'c}}, \bibinfo{person}{Adri{\`a}~Puigdom{\`e}nech Badia}, \bibinfo{person}{David Budden}, \bibinfo{person}{Razvan Pascanu}, \bibinfo{person}{Andrea Banino}, \bibinfo{person}{Misha Dashevskiy}, \bibinfo{person}{Raia Hadsell}, {and} \bibinfo{person}{Charles Blundell}.} \bibinfo{year}{2022}\natexlab{}.
\newblock \showarticletitle{The CLRS algorithmic reasoning benchmark}. In \bibinfo{booktitle}{\emph{International Conference on Machine Learning}}. PMLR, \bibinfo{pages}{22084--22102}.
\newblock


\bibitem[\protect\citeauthoryear{White, Safari, Sukthanker, Ru, Elsken, Zela, Dey, and Hutter}{White et~al\mbox{.}}{2023}]%
        {white2023neural}
\bibfield{author}{\bibinfo{person}{Colin White}, \bibinfo{person}{Mahmoud Safari}, \bibinfo{person}{Rhea Sukthanker}, \bibinfo{person}{Binxin Ru}, \bibinfo{person}{Thomas Elsken}, \bibinfo{person}{Arber Zela}, \bibinfo{person}{Debadeepta Dey}, {and} \bibinfo{person}{Frank Hutter}.} \bibinfo{year}{2023}\natexlab{}.
\newblock \showarticletitle{Neural architecture search: Insights from 1000 papers}.
\newblock \bibinfo{journal}{\emph{arXiv preprint arXiv:2301.08727}} (\bibinfo{year}{2023}).
\newblock


\bibitem[\protect\citeauthoryear{Wistuba}{Wistuba}{2017}]%
        {wistuba2017finding}
\bibfield{author}{\bibinfo{person}{Martin Wistuba}.} \bibinfo{year}{2017}\natexlab{}.
\newblock \showarticletitle{Finding competitive network architectures within a day using uct}.
\newblock \bibinfo{journal}{\emph{arXiv preprint arXiv:1712.07420}} (\bibinfo{year}{2017}).
\newblock


\bibitem[\protect\citeauthoryear{Ying, Klein, Christiansen, Real, Murphy, and Hutter}{Ying et~al\mbox{.}}{2019}]%
        {ying2019bench}
\bibfield{author}{\bibinfo{person}{Chris Ying}, \bibinfo{person}{Aaron Klein}, \bibinfo{person}{Eric Christiansen}, \bibinfo{person}{Esteban Real}, \bibinfo{person}{Kevin Murphy}, {and} \bibinfo{person}{Frank Hutter}.} \bibinfo{year}{2019}\natexlab{}.
\newblock \showarticletitle{Nas-bench-101: Towards reproducible neural architecture search}. In \bibinfo{booktitle}{\emph{International Conference on Machine Learning}}. PMLR, \bibinfo{pages}{7105--7114}.
\newblock


\bibitem[\protect\citeauthoryear{Zheng, Su, You, Wang, Qian, Xu, and Albanie}{Zheng et~al\mbox{.}}{2023}]%
        {zheng2023can}
\bibfield{author}{\bibinfo{person}{Mingkai Zheng}, \bibinfo{person}{Xiu Su}, \bibinfo{person}{Shan You}, \bibinfo{person}{Fei Wang}, \bibinfo{person}{Chen Qian}, \bibinfo{person}{Chang Xu}, {and} \bibinfo{person}{Samuel Albanie}.} \bibinfo{year}{2023}\natexlab{}.
\newblock \showarticletitle{Can GPT-4 Perform Neural Architecture Search?}
\newblock \bibinfo{journal}{\emph{arXiv preprint arXiv:2304.10970}} (\bibinfo{year}{2023}).
\newblock


\end{thebibliography}


\end{document}